\documentclass[runningheads]{llncs}
\usepackage{graphicx}
\usepackage{comment}
\usepackage{amsmath,amssymb} 
\usepackage{color}

\usepackage{times}
\usepackage{epsfig} 
\usepackage{graphicx}
\usepackage{amsmath}
\usepackage{amssymb}

\usepackage[utf8]{inputenc} 
\usepackage[T1]{fontenc}    
\usepackage{url}            
\usepackage{booktabs}       
\usepackage{amsfonts}       
\usepackage{nicefrac}       
\usepackage{microtype}      

\usepackage{multicol}

\usepackage{booktabs}

\usepackage[pagebackref=true,breaklinks=true,colorlinks,bookmarks=false]{hyperref}

\newcommand{\ask}[1]{#1}

\newcommand{\review}[1]{\textcolor{black}{#1}}

\def\vec#1{\mathchoice{\mbox{\boldmath$\displaystyle#1$}}
	{\mbox{\boldmath$\textstyle#1$}}
	{\mbox{\boldmath$\scriptstyle#1$}}
	{\mbox{\boldmath$\scriptscriptstyle#1$}}}

\newcommand{\class}{c}
\newcommand{\x}{\vec x}
\newcommand{\vv}{\vec v}
\newcommand{\DR}{\ensuremath{\text{DR}}}
\newcommand{\ResnetES }{ResNet-18 }
\newcommand{\ResnetVS }{ResNet-34 }
\newcommand{\ResnetS}{ResNet-50 }
\newcommand{\ResnetB}{ResNet-101 }
\newcommand{\ResnextS}{ResNeXt-101 (32$\times$8$d$) }
\newcommand{\ResnextSshort}{ResNeXt-101 }
\newcommand{\SENet}{SENet-154 }
\newcommand{\inceptionresnet}{InceptionResNetV2 }

\newcommand{\WRNshort}{WRN }
\newcommand{\WRNnoblank}{WRN (28-10-dropout)}
\newcommand{\DenseNet}{DenseNet-BC ($k$ = 40, depth=190) }
\newcommand{\DenseNetshort}{DenseNet-BC }

\newcommand{\cifar}{CIFAR100}
\newcommand{\DML}{\ensuremath{\text{DML}}}
\newcommand{\DMLone}{{\DML$_1$}}
\newcommand{\DMLtwo}{{\DML$_2$}}

\begin{document}
\pagestyle{headings}
\mainmatter
\def\ECCVSubNumber{6624}  

\title{Label-similarity Curriculum Learning} 

\titlerunning{Label-similarity Curriculum Learning}
%
\author{{\"U}r{\"u}n Dogan\inst{1} \and
Aniket Anand Deshmukh\inst{1} \and
Marcin Bronislaw Machura\inst{1} \and Christian~Igel\inst{2}}
%
\authorrunning{{\"U}. Dogan, A. Deshmukh,  M. Machura, C. Igel}
%
\institute{Microsoft Ads, Microsoft, Sunnyvale, CA, USA \\
\email{udogan@microsoft.com, andeshm@microsoft.com, mmachura@microsoft.com} \and
Department of Computer Science, University of Copenhagen, Denmark\\ 
\email{igel@di.ku.dk}}
\maketitle

\begin{abstract}
Curriculum learning can improve neural network training by guiding
  the optimization to desirable optima.  We propose a novel
  curriculum learning approach for image classification that adapts
  the loss function by changing the label representation.
  
  The idea is to use a probability distribution over classes as target label, where the class probabilities reflect the similarity to the true class.  Gradually, this label representation is shifted towards the standard one-hot-encoding.
  That is, in the beginning minor mistakes are corrected less than large
  mistakes, resembling a teaching process in which broad concepts are
  explained first before subtle differences are taught.
   
 The class similarity can be based on  prior knowledge.
 For the special case of the labels being  natural words, we propose a generic way to automatically compute the similarities. The natural words are  embedded into Euclidean space using a standard word embedding.  The probability of each class is then a function of the cosine similarity between the vector representations of the class and the true label.

  The proposed label-similarity curriculum learning (LCL) approach was empirically evaluated using several popular deep learning architectures for image classification tasks applied to five datasets including ImageNet, CIFAR100, and AWA2.
  In all scenarios, LCL was able to improve the
  classification accuracy on the test data compared to standard training. Code to reproduce results is available at \href{https://github.com/speedystream/LCL}{https://github.com/speedystream/LCL}. 
 
\end{abstract}


\section{Introduction}
\label{sec:introduction}
When educating humans, the teaching material is typically presented
with increasing difficulty. \emph{Curriculum learning} adopts this
principle for machine learning to guide an iterative optimization
method to a desirable optimum. In curriculum learning for neural
networks as proposed by Bengio et al.~\cite{bengio2009curriculum}, the
training examples are weighted. In the beginning of the training, more
weight is put on ``easier'' examples. The weighting is gradually
changed to  uniform weights corresponding to the canonical
objective function.

\review{Inspired by Bengio et al.~\cite{bengio2009curriculum},}
we propose \emph{label-similarity curriculum learning} (LCL) as another way
to  ``learn easier aspects of the task or easier sub-tasks, and then gradually increase the difficulty level.''
%
If a toddler who is just learning to speak points at a car and utters
``cow'', a parent will typically react with some teaching signal.
However, a young infant is not expected 
to discriminate between a cheetah and a leopard, and mixing up the two
would only lead to a very mild correction signal -- if at all.
With increasing age, smaller errors will also be communicated.

We transfer this approach to neural network training for classification
tasks. 
Instead of a one-hot-encoding, the target represents a probability distribution
over all possible classes. The probability of each class depends on
the similarity between the class and the true label. That is, instead
of solely belonging to its true  class, each input can also
belong to similar classes to a lesser extent.  Gradually, this label representation is shifted towards the standard one-hot-encoding, where  targets representing different classes are orthogonal. In the beginning of training, the targets of inputs with labels \texttt{cheetah} and \texttt{leopard} should almost be the same, but always be very different from \texttt{car}. During the training process, the label representation is gradually morphed into the one-hot encoding, decreasing the entropy of the distribution encoded by the target over time.  That is, in the beginning small mistakes -- in the sense that similar classes are mixed up -- are corrected less than big mistakes, resembling a teaching process in which broad concepts are
explained first before subtle differences are taught.

The question arises how to define a proper similarity between classes. One can get a label-similarity matrix based on prior knowledge or some known structure.  For the case where the similarity is not explicitly given and the labels correspond to natural language words, we propose a way to automatically infer a representation that reflects semantic similarity. We map the labels into a Euclidean space using a word embedding. Concretely, this is done  by applying a generic document embedding to a  document explaining the label (its Wikipedia entry). Then the cosine similarities between the vector representations of the label and all possible classes are  computed. Based on these values,  a distribution over the possible classes is defined which serves as the learning target. 

%

Our way to define the target representation
resembles the idea of \emph{hierarchical loss functions}
\cite{Silla2011,DBLP:journals/corr/abs-1709-01062}  (``We define a
metric that, inter alia, can penalize failure to distinguish between a
sheepdog and a skyscraper more than failure to distinguish between a
sheepdog and a poodle.'' \cite{DBLP:journals/corr/abs-1709-01062}).
However, there are two decisive differences. First,
we propose to gradually  shift from a ``hierarchical loss'' to
a ``flat loss''.
Second,  unlike in \cite{DBLP:journals/corr/abs-1709-01062}, our approach does not necessarily presume a given hierarchy. When dealing with natural language labels, we propose a way to automatically infer the
 similarity from a generic word embedding 
under the assumption that exploiting  semantic
similarity can be helpful in guiding the learning process.

For evaluating  label-similarity curriculum learning (LCL), we need data with some structure in the label space that curriculum learning can exploit. Furthermore, there should be sufficiently many classes and the task should not be easy to learn. To get label similarity based on word embeddings, we need a dataset with  natural language labels. In this study, we focus on three popular benchmark datasets, ImageNet \cite{deng2009imagenet}, CIFAR100  \cite{krizhevsky2009learning}, and Animals with Attributes (AwA)  \cite{xian2018zero}.
{To show the generality of our approach, we consider different deep learning architectures, and also different preprocessing and learning processes.}
The time schedule for increasing the ``difficulty'' of the learning
task is an obvious hyperparameter, which we carefully study \ask{and show to have little importance}.

The next section points to related literature and  Section \ref{sec:proposed_method} introduces the new \emph{label-similarity curriculum learning}.
Section~\ref{sec:experiments} describes the experiments and
Section~\ref{sec:results} the results before we conclude.

\section{Related Work}
\label{sec:related_work}
Starting from the work by  Bengio et  al.~\cite{bengio2009curriculum}, a variety of curriculum learning  approaches has been
studied. However, they all define a curriculum at the level of
training examples. For instance, \emph{self-paced learning} by 
Kumar et al.~\cite{kumar2010self}
%
introduces 
latent variables for modelling ``easiness'' of an examples.
Graves et al.~\cite{graves2017automated} consider example-based
improvement measures as reward signals for multi-armed bandits, which
then build stochastic syllabi for neural networks.
\review{
Florensa et al.~\cite{florensa2017reverse} study curriculum learning
in the context of reinforcement learning in robotics. They propose to
train a robot by gradually increasing the complexity of the task at
hand (e.g., the  robot learns to reach a goal by setting starting points increasingly far from the goal).} 
\review{
In recent work, Weinshall et al.~\cite{weinshall2018curriculum}
consider learning tasks with convex linear regression loss and prove
that the convergence rate of a
perfect curriculum learning method increases with the difficulty of the examples. In addition, they propose a  method which infers the curriculum
using transfer learning from another network (e.g., \ResnetS)
pretrained on a different task. They train a linear classifier using
features extracted from the pretrained model and score each training
example using the linear classifier’s confidence (e.g., the
margin of an SVM).
Finally, they train a smaller deep neural network for the transfer
learning task following a curriculum based on these scores.} 

Buciluǎ et al.~\cite{bucilua2006model} have  proposed compressing a large model into a simple model which  reduces space requirements and increases inference speed at the cost of a small performance loss. This idea has been revisited in  \cite{hinton2015distilling} under the name  \emph{knowledge distillation} (KD) and received a significant amount of attention (e.g., \cite{papernot2016distillation,romero2014fitnets,zagoruyko2016paying,yim2017gift,orbes-ortega:19b}).
KD  methods typically require a pretrained model to start with or train a series of models on the same training data.
Standard KD considers a teacher network and a student network. 
The 
powerful teacher network is used to support the  training  of the student network which may be less complex or may have access to less data for training. KD is related to curriculum learning methods because the teacher network guides the learning of student networks \cite{hinton2015distilling}. A variant of KD,  \emph{born again neural network}, trains a series of models, not only one \cite{furlanello2018born}. 

\emph{Deep mutual learning} (DML) is also loosely related to our proposed approach \cite{hinton2015distilling,zhang2018deep}.  In DML, two models  solve the same classification problem collaboratively and are jointly optimised   \cite{zhang2018deep}. Each model acts as a teacher for the other model, and each network is trained with two losses. The first loss is the standard cross-entropy  between the model's predictions and target labels. The second  is a \emph{mimicry loss} that aligns both model's class posteriors with the class probabilities of the respective other  model. 

Another related approach is \emph{CurriculumNet} \cite{guo2018curriculumnet}, a clustering based curriculum strategy for learning from noisy data. CurriculumNet consists of three steps. First, a deep neural network is trained on the noisy label data. Second, features are extracted by using the model trained in the first step. Using clustering algorithms, these features are then grouped into different sets and sorted into easy and difficult examples. Finally, a new deep neural network is trained using example-weighted curriculum learning. Sorting of examples from easy to hard and clustering algorithms add many hyper-parameters (e.g., number of clusters), and one has to train two neural network models of almost the same size.

Our algorithm can be considered as a multi-modal deep learning method,  where text data is used for estimating the class similarity matrix to improve image classification. 
However, it is different from standard multimodal methods as it does not use text data as an input to the deep neural network. The \emph{DeVise} algorithm is a popular multi-modal method which utilizes the text modality in order to learn a mapping from an image classifier's feature space to a semantic space.   \cite{frome2013devise}. {DeVise} requires a pretrained deep neural network. Furthermore, as  stated in \cite{frome2013devise}, it does not improve the accuracy on the original task but  aims at training a  model for zero-shot learning.

There is an obvious relation between LCL and \emph{label smoothing} (LS) \cite{muller2019does}, which we will discuss in Section~\ref{sec:experiments}. 

The computational requirements of KD, DML,  and CurriculumNet  are significantly higher compared to our method, which is rather simple. Furthermore, our method does not require training more than one model and adds only a single  hyper-parameter. 

\section{Method}
\label{sec:proposed_method}

We assume a discrete set of training examples $(\x_1, \class_1)
,\dots,(\x_\ell, \class_\ell)  \in \mathcal X\times \mathcal C$,
with input space  $\mathcal X$ and finite label space $\mathcal C$ with
cardinality $|\mathcal C| = C$.
Let $n: \mathcal C\to \{1,\dots,C\}$ be a bijective mapping assigning 
each label to a unique integer. This allows a straight-forward
definition of the  one-hot encoding $\vec y_i\in\mathbb R^C$
for each  training example   $(\x_i, \class_i)$. The $j$-th component of $\vec
y_i$, which
is denoted by $[\vec y_i]_j$, equals 1 if \ask{$n(\class_i)=j$} and 0 otherwise.
\begin{figure*}
	\includegraphics[width=\linewidth]{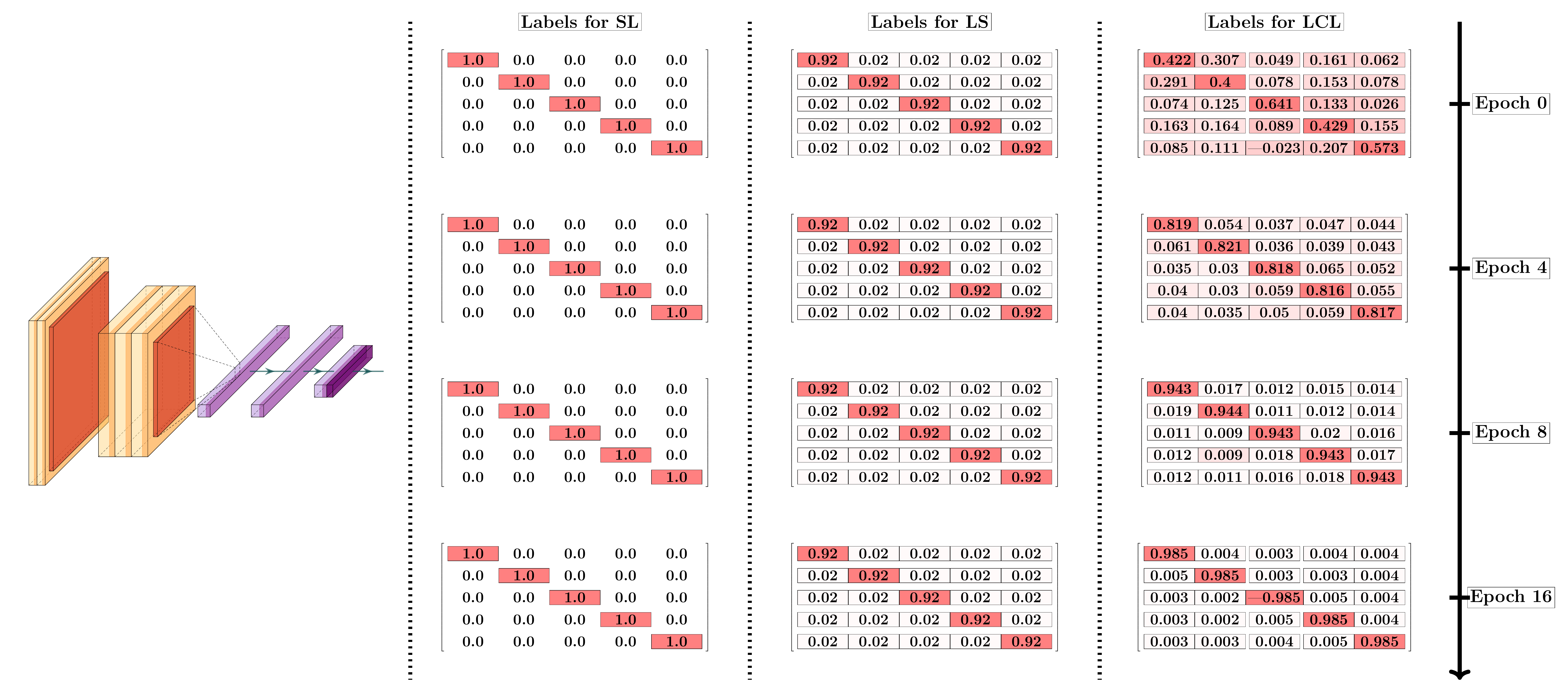}
	\caption{A deep network (left) trained with three different encodings  on a  five-class dataset  with labels Lion, 
	Tiger, Aircraft Carrier, Alaskan Wolf and Mushroom. 
	The SL (standard learning,  see Section~\ref{sec:experiments} for details) column shows the label matrix for one-hot-encoding.
	When  using LS (label smoothing, see Section~\ref{sec:experiments}),  the loss between the network output and a 
	smoothed version of the label, which does not change over time, is minimized. We propose to use a probability distribution over classes as target label, 
	where the class probabilities reflect the similarity to the true class. This is shown in the  LCL column. Unlike LS the proposed label encoding changes during training and converges 
	to the original optimization problem solved when using SL.}
	\label{fig:comparative_model_exp}
\end{figure*}

\subsection{Document embedding for defining label similarity}
\label{subsec:semantic_similarity}
Our learning curriculum is based on the pairwise similarities between the $C$
 classes, which are defined  based on
the semantic similarity of the class labels.
Now assume that the labels are 
natural language words, for example
$\mathcal C=\{  \dots$, \texttt{flute}, $\dots$, \texttt{strawberry}, $\dots$, \texttt{backpack},
$\dots\}$.
To quantify semantic similarity, we embed the natural language labels into Euclidean
space using a word embedding \cite{DBLP:journals/corr/abs-1301-3781}
such that similar words are nearby in the new representation.

\begin{figure}
\centering
	\includegraphics[width=0.5\linewidth]{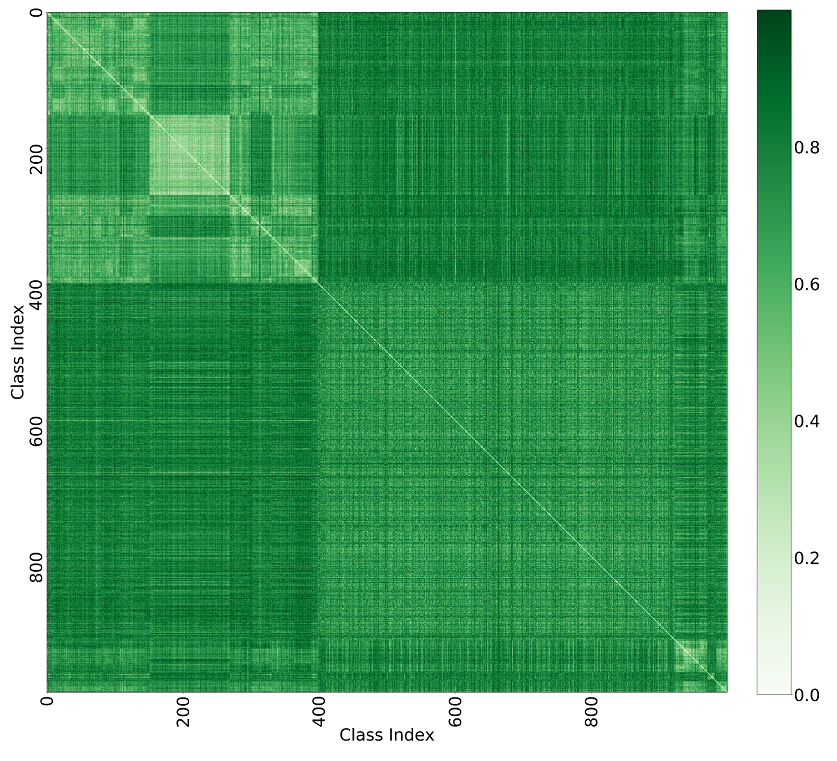}
	\caption{Label dissimilarity matrix visualizing $1-s(c_i,c_j)$, $1\le
          i,j\le 1000$, for ImageNet.}
	\label{fig:class_dis_matrix}
\end{figure}

ImageNet labels are given by WordNet identifiers representing synsets, and we redefine the labels for other datasets in a similar way. First, we convert synsest to words, for example, \texttt{n02119789} to
``fox''. Then, we find the Wikipedia article describing each word, for instance, ``Orange (fruit)'' was selected for \texttt{orange}.
Then we apply \texttt{doc2vec} \cite{le2014distributed} for mapping
the article into Euclidean space.
We used a generic   \texttt{doc2vec}  embedding trained on the
English  Wikipedia corpus.
This gives us the  encoding $f_{\text{enc}}:\mathcal C\to\mathbb R^d$,
mapping each class label to the corresponding Wikipedia article and
then computing the corresponding vector representation using
\texttt{doc2vec} (with $d=100$, see below).
Now we can compute the
similarity between two classes $c_i$ and $c_j$ by the cosine
similarity
\begin{equation}\label{eq:sim}
s(c_i,c_j)=\frac{\langle f_{\text{enc}}(c_i), f_{\text{enc}}(c_j)
  \rangle}{\| f_{\text{enc}}(c_i)\|\|f_{\text{enc}}(c_j)\|} \enspace,
\end{equation}
which in our setting is  always non-negative.
The resulting label dissimilarity matrix for the ImageNet labels is visualized in Figure~\ref{fig:class_dis_matrix}. 

\subsection{Label encoding}

We adopt the formal definition of a curriculum
from
the seminal paper by Bengio et al.~\cite{bengio2009curriculum}.
In  \cite{bengio2009curriculum}, 
a weighting of the training data is adapted, so that in the
beginning a larger weight is put on easy examples.
To distinguish this work from our approach, we refer to 
it as \emph{example-weighting curriculum}.

Let $t\ge 0$ denote some notion of  training time (e.g., a counter of
training epochs).
In \cite{bengio2009curriculum}, there is a sequence of weights
associated with each example  $i=1,\dots,\ell$, which we denote by
$w_i^{(t)}\in [0,1]$.
These weights are normalized so that $\sum_{i=1}^{\ell}w^{(t)}_i=1$ to 
describe a proper probability distribution over the training examples.

For the weight sequence to be a proper (example-weighting) curriculum,
Bengio et al.~\cite{bengio2009curriculum} demand that the entropy of
the weights
\begin{equation}
H(\vec w^{(t)})=-\sum_{i=1}^\ell  w_i^{(t)} \ln w_i^{(t)}
\end{equation}
is monotonically increasing with $t$ (the weights should converge
to the uniform distribution).

We define our \emph{label-weighting curriculum}
in a similar axiomatic way. Instead of a sequence of weights for the
training examples varying with $t$, we have a sequence of label
vectors for each training example. Let $\vv_i^{(t)}$ 
denote the $C$-dimensional label vector for training pattern $i$ at time $t$.
For the sequence to be a  {label-weighting curriculum},
the entropy of the label vector components 
\begin{equation}
\forall i=1,\dots,\ell: H(\vec \vv_i^{(t)})=-\sum_{c=1}^C  [{\vec
	\vv}_i]_c^{(t)} \ln [{\vec \vv}_i]_c^{(t)}
\end{equation}
should be monotonically 
\emph{decreasing} under the constraints that for each label vector $\vec \vv_i^{(t)}$ we
have $[{\vec \vv}]_j\geq 0$ for all $j$, $\|\vec \vv_i^{(t)}\|_1=1$, and
$\operatorname{argmax}_j [{\vec \vv}]_j^{(t)} = n(c_i)$ for all $t$.
The conditions imply that $\vv_i$ is always an element of the
probability simplex, the class label given in the training set
always gets the highest probability, and $\vec \vv_i^{(t)}$
converges to $\vec y_i$.


We now give an example of how to adapt the label vectors. 
Similar as in \cite{bengio2009curriculum},
we define for each training example $i$
the simple update rule:
\begin{equation}\label{eq:eps}
[\vec \vv_i]_j^{(t+1)} =
\begin{cases}
\frac{1}{1 + \epsilon \sum_{ k \ne n(c_i) } [\vec \vv_i]_k^{(t)} }  & \text{if $j= n(\class_i)$}\\
\frac{\epsilon  [\vec \vv_i]_j^{(t)}}{1 + \epsilon \sum_{ k \ne n(c_i) } [\vec \vv_i]_k^{(t)} }   & \text{otherwise}\\
\end{cases}
\end{equation}
The constant parameter $0<\epsilon<1$ controls how quickly the label vectors converge to the one-hot-encoded labels.
This update rule  leads to a proper {label-weighting curriculum}.
During learning, the entries for all components except $n(\class_i)$ drop with
$\mathcal O(\epsilon^t)$. 
Note that $[\vec \vv_i]_{n(c_i)}^{(t+1)} \ge [\vec \vv_i]_{n(c_i)}^{(t)}$.
The  vectors are initialized using the label similarity defined in \eqref{eq:sim}:
\begin{equation}
  [{\vec \vv}_i]_j^{(0)} = 
  \frac{s\big( c_i, n^{-1}(j) \big)}{\sum_{k=1}^{C} s\big( c_i, n^{-1}(k) \big) }
\end{equation}
Recall that $n^{-1}(j) $ denotes  the ``$j$-th'' natural language class label.


\subsection{Loss function}
Let $\mathcal{L}$ be a loss function between two probability distributions and $ f_{\theta}(x)$ be the predicted distribution for example $\vec x$ for some model parameters $ \theta$. At time step $t$ we optimize
$J^{(t)}(\theta) = \sum_{i = 1}^n \mathcal{L} (f_{\theta}(\vec{x}_i), \vec \vv_i^{(t)}) + \lambda r(\theta)$,
where $\lambda$ is a positive constant and $r(\theta)$ is a regularization function. In this paper, the networks are trained using the standard cross-entropy loss function with normalized targets $\vec v_i$ for the inputs $\x_i$, $i=1,\dots,\ell$. Hence, in the beginning, predicting the 
correct one-hot encoded label $\vec y_i$ causes an error signal. That is, initially it is less penalized if an object is not correctly classified with maximum confidence. Later in the training process, $\vec v_i$  converges to $\vec y_i$ and  the classifier  is then pushed to build up confidence.

\section{Experiments}
\label{sec:experiments}
We evaluated our curriculum  learning strategy by running extensive experiments on ImageNet~\cite{deng2009imagenet}, CIFAR$100$~\cite{krizhevsky2009learning}, and AWA$2$~\cite{xian2018zero} plus additional experiments on CUB-200-2011 \cite{WahCUB_200_2011} and NABirds \cite{Horn_2015_CVPR} (see supplementary material).
On CUB-200-2011 and AwA2, we evaluated our approach using both the proposed semantic similarity of the labels as well as visual similarity. On NABirds, we evaluated our approach also using similarity based on the given (biological) hierarchy, where we used simrank~\cite{jeh2002simrank} for calculating the similarity matrix.

For each dataset we considered at least two different models and two different baselines. 
Descriptive statistics of the datasets and a summary of the experimental setup are given in Table \ref{tab:data_statistics} and Table \ref{tab:all_exp}. 
We considered different training set sizes, where $\DR\in\{5\%,10\%,20\%,100\%\}$ refers to the fraction of training data used. The remaining training data was discarded (i.e., not used in the training process at all); the test data were always the same. 

\begin{table}[ht]
\begin{center}
\caption{$\ell_{\text{train}}$ denotes the number of training images, $\ell_{\text{test}}$ denotes the number of test images and  $C$ the number of classes in a given dataset;
DR indicates the data set sizes (percentage of $\ell_{\text{train}}$), $\epsilon$  the cooling parameters, and \#Rep  the number of repetitions with different initializations/seeds. The column Sim.{} indicates which similarity measures were used, where l stands for the semantic similarity using the word embedding of the labels, v for a  measure based on the similarity of the images, and h for similarity based on a given label hierarchy.}
\label{tab:data_statistics}
\begin{tabular}{lrrr@{\qquad}cccc}
& $\ell_{\text{train}}$ & $\ell_{\text{test}}$ & $C$ & \textbf{DR} & \textbf{\#Rep} & $\boldsymbol{\epsilon}$ & \textbf{Sim.}\\
\midrule
AWA2     &  29865   &  7457   &  50  & $5\%$, $10\%$, $20\%$, $100\%$ & 4     & 0.9, 0.99, 0.999 & l, v \\
CIFAR100 &   50000  &   10000  &100 & $5\%$, $10\%$, $20\%$, $100\%$ & 4     & 0.9, 0.99, 0.999 & l    \\
ImageNet &   1281167  & 50000    & 1000 & $5\%$, $10\%$, $20\%$, $100\%$ & 4     & 0.9, 0.99, 0.999 & l\\   
NABirds  &23912&24615&555&100\%&4&0.9,0.99,.999& l, h\\
CUB-200-2011  &5994&5794&201&100\%&4&0.9,0.99,.999& l, v\\
\bottomrule
\end{tabular}
\end{center}
\end{table}

We  empirically compared  the following algorithms:
\begin{enumerate}
    \setlength\itemsep{-0.2em}
        \item Label-similarity curriculum learning (LCL): Proposed method with label update rule \eqref{eq:eps}. The time step $t$ is the epoch number. 
        \item Standard Learning (SL): This is a standard  setup  with fixed one-hot encoding.
    \item Label Smoothing (LS): Label smoothing uses soft targets instead of one-hot encoding. It has been argued that LS prevents the network from becoming over-confident and improves the empirical performance of the algorithm \cite{muller2019does}. For  $ 0 \leq \alpha \leq 1 $ label smoothing uses following label vector 
            \begin{equation}\label{eq:LS}
                [\vec \vv_i]_j^{(t)} =
                \begin{cases}
                (1 - \alpha) + \frac{\alpha}{C} & \text{if $j= n(\class_i)$}\\
                 \frac{\alpha}{C}  & \text{otherwise}\\
                \end{cases}\enspace\enspace \text{for all } t.
            \end{equation} 
     We set$ \alpha = 0.1 $ for the evaluations in this study.        
    \item Deep Mutual Learning (DML):  In DML, two models, referred to as \DMLone and \DMLtwo,  solve the same classification problem collaboratively and are  optimised jointly  \cite{zhang2018deep}. It uses one hot-encoding along with cross-entropy loss as in SL but adds additional terms 
    $\text{KL}( \hat{\vec \vv}_{\text{DML}_1}^{(t)}\,\|\, \hat{\vec \vv}_{\text{DML}_2}^{(t)} ) + \text{KL}( \hat{\vec \vv}_{\text{DML}_2}^{(t)}\,\|\,  \hat{\vec \vv}_{\text{DML}_1}^{(t)} )$,
    where $\text{KL}$ denotes the  Kullback–Leibler  divergence
    and $\hat{\vec \vv}_{\text{DML}_1}^{(t)} $ 
    and $ \hat{\vec \vv}_{\text{DML}_2}^{(t)} $  are the predicted label probability vectors for both models. 
    We report the classification performance of both \DMLone\ and \DMLtwo.
    \item Knowledge Distillation (KD): In KD, one model is trained first using one-hot encoded targets, and then the class probabilities produced by the first model are used as ``soft targets'' for training the
    second model \cite{hinton2015distilling}.
    \item Curriculum Net (CN): In CN, example-weighted curriculum is built by sorting examples from easy to hard  \cite{guo2018curriculumnet}. 
\end{enumerate}

\begin{table*}[ht]
\begin{center}
\caption{ \ResnextSshort denotes \ResnextS, \WRNshort denotes \WRNnoblank, and \DenseNetshort denotes \DenseNet.
}
\label{tab:all_exp}
\begin{tabular}{l@{\quad}c@{\quad}c}
\textbf{Model} & \textbf{Dataset}  & \textbf{Baselines} \\
\toprule  
\ResnetES~\cite{xie2017aggregated} & CUB-200-2011  & LS, DML, SL, KD, CN \\
\ResnetVS~\cite{xie2017aggregated} & CUB-200-2011, NABirds & LS, DML, SL, KD, CN \\
\ResnetS~\cite{he2016deep} & ImageNet, NABirds & SL, LS, KD, CN  \\
\ResnextSshort~\cite{xie2017aggregated} & ImageNet & SL, LS, KD, CN   \\
\SENet~\cite{hu2018squeeze}& ImageNet & SL, LS, KD, CN   \\
\ResnetB~\cite{he2016deep} & AWA2     & LS, DML, SL, KD, CN \\
\inceptionresnet~\cite{szegedy2017inception}& AWA2     & LS, DML, SL, KD, CN \\
\WRNshort~\cite{zagoruyko2016wide} & CIFAR100 & LS, DML, SL, KD, CN \\
\DenseNetshort~\cite{huang2017densely} & CIFAR100 & LS, DML, SL, KD, CN \\
\bottomrule
\end{tabular}
\end{center}
\end{table*}
For all architectures, we have followed the experimental protocols described in the original publications \cite{he2016deep,zagoruyko2016wide,xie2017aggregated,hu2018squeeze,szegedy2017inception,huang2017densely}. 
All experiments were conducted using the PyTorch deep learning library \cite{paszke2017automatic}.\footnote{The code to reproduce our results is available in the supplementary material.} For all experiments, except ImageNet $\DR=100\%$, we  used stochastic gradient descent (SGD) for optimization. For ImageNet $\DR=100\%$  we used the distributed SGD algorithm \cite{goyal2017accurate} with Horovod\footnote{Horovod is a method which uses large batches over multiple GPU nodes and some accuracy loss is expected  for the baseline method and this is well established. For more details please see Table~$1$ and Table $2$.c in  \cite{paszke2017automatic}.} \cite{sergeev2018horovod} support because of the computational demands.
The distributed SGD algorithm \cite{goyal2017accurate} is one of the state-of-the-art methods for large scale training. It is expected to lead to a slight loss in performance when a large batch size is used (see \cite{goyal2017accurate} for details). 

Our approach introduces the hyperparameter $\epsilon$, see \eqref{eq:eps}. 
In order to assess the stability of the proposed method, we present results for $\epsilon \in \{0.9,0.99,0.999\}$.\footnote{We have tried $\epsilon \in \{0.8, 0.9, 0.91, \ldots,$ $0.98, 0.99, 0.992, \ldots,$ $0.998, 0.999\}$ for \ResnetS\ and \ResnetB. The results showed that the search space for $\epsilon$ can be less granular and we have limited the search space accordingly.} We repeated all experiments four times. We report the top-$1$ and top-$5$ classification accuracy on the \ask{test} datasets (standard deviations are  reported in the supplementary material). 


For estimating the label similarity matrix, we  used \emph{pretrained} \texttt{doc2vec} emebeddings with dimensions $d\in\{100, 300, 500\}$ with \ResnetS\ and \ResnetB. We did not observe any significant differences in the classification accuracies. The maximum difference between compatible settings were less than 0.06 \%. Hence, we only report results for  the $d=100$ dimensional \texttt{doc2vec} embeddings.

For each experiment, we used workstations having 4 Tesla P100 GPUs (with 16GB GPU RAM) each. For network communication we used InfiniBand, which is a computer-networking communications standard designed for high throughput and low-latency scenarios. 

We tuned the hyperparameters for the baseline method (SL) only. For 100\% data, we took the hyperparameters from the original publications. For all other settings, we optimized learning rate, batch size and weight-decay for the standard baseline (SL). Then we use the very same parameters for our approach (we just varied the new parameter epsilon). Thus, hyperparameter tuning would rather increase the gain from using our method.
Thus, one might argue that the new algorithm is using sub-optimal hyperparameters compared to the baselines. However, our goal was to  show that the proposed algorithm can improve any model on different datasets without tuning hyperparameters.

\section{Results and Discussion}
\label{sec:results}
\begin{figure}[t!]
	\includegraphics[width=.32\linewidth]{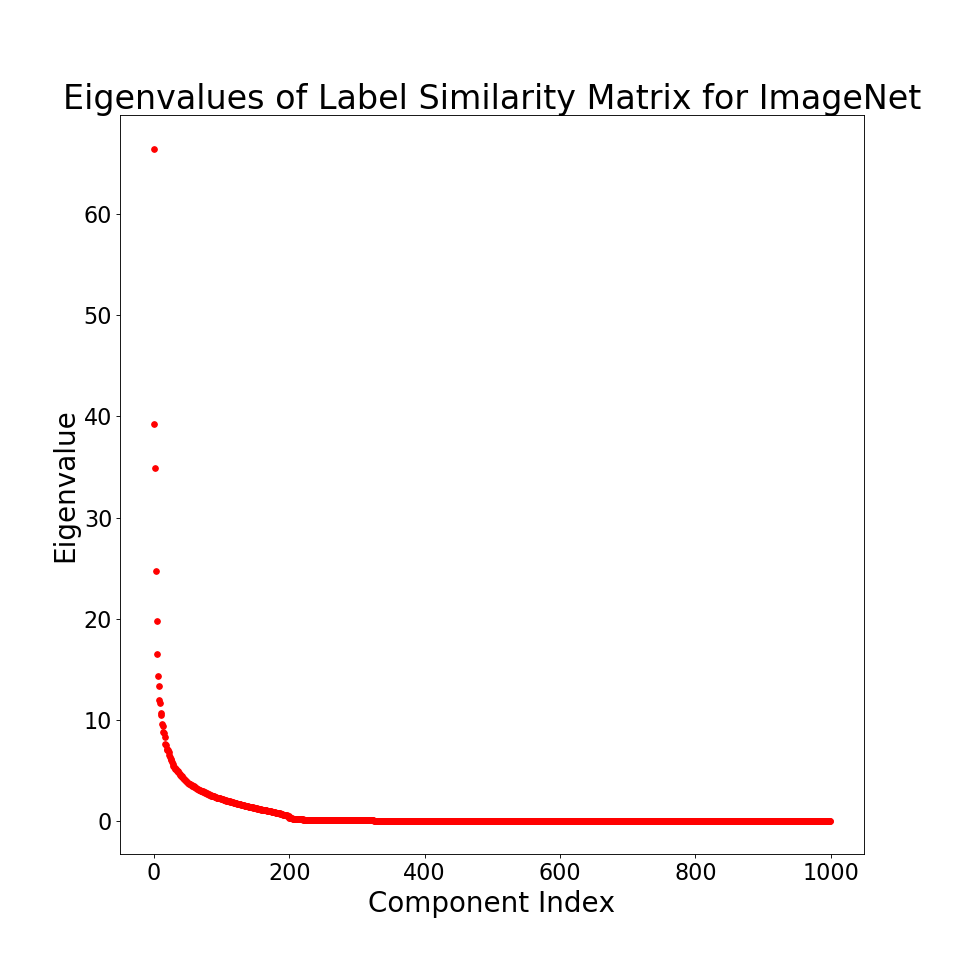}
	\includegraphics[width=.32\linewidth]{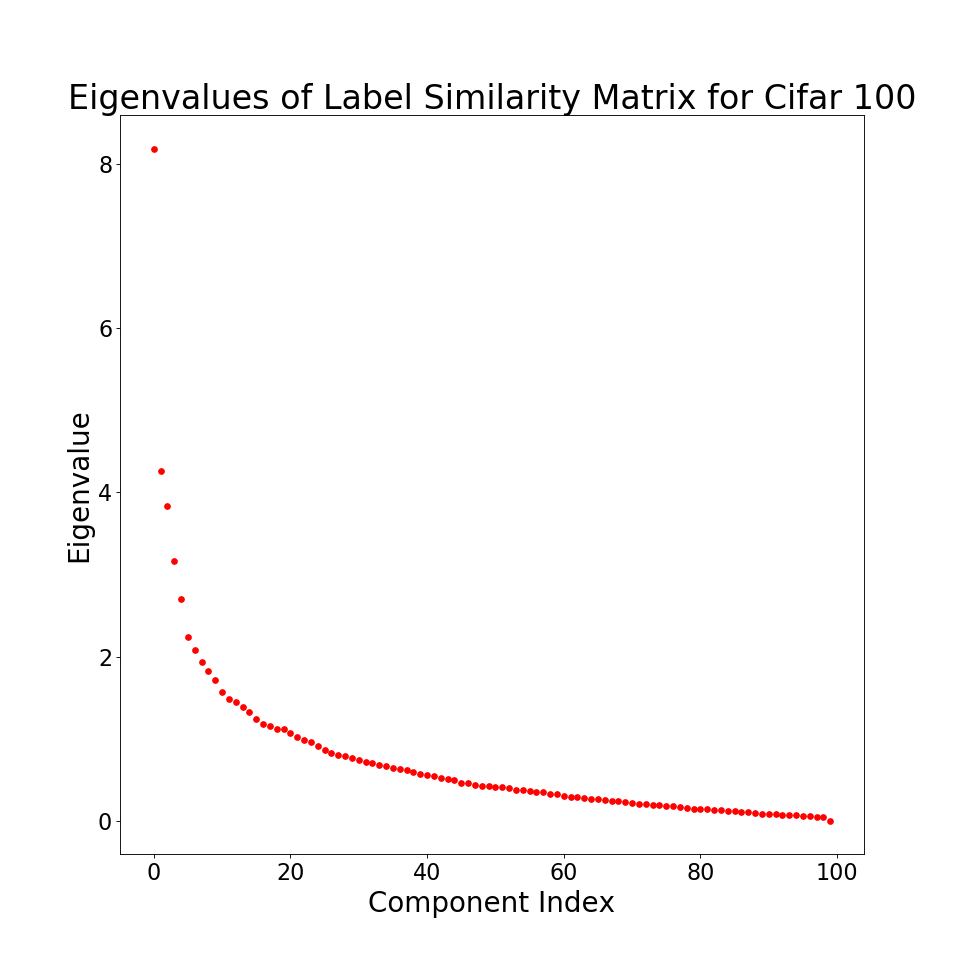}
	\includegraphics[width=.32\linewidth]{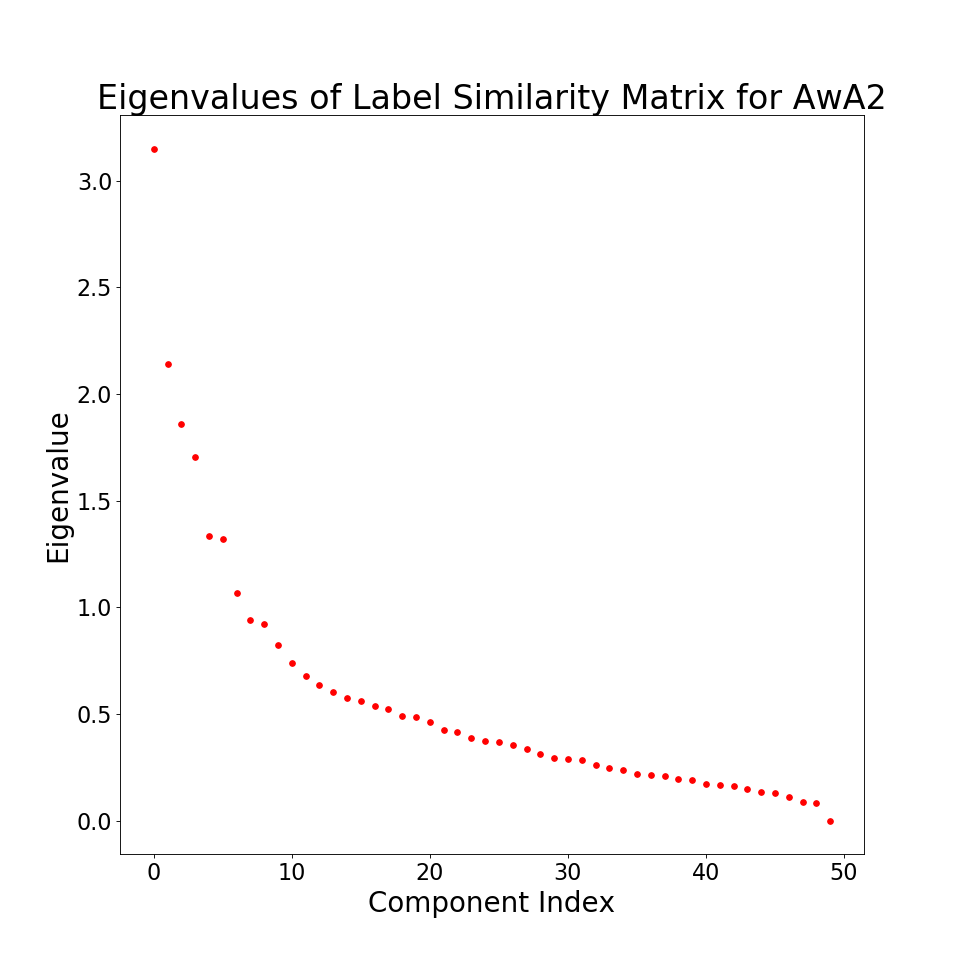}
	\caption{Eigenvalue distributions of the class similarity matrices for ImageNet, \cifar{}, and AwA2.}
	\label{fig:eig_imagenet_class_dis_matrix}
\end{figure}

We will focus on the results for ImageNet, CIFAR100, and AWA2 and the similarity measure introduced in Section~\ref{sec:proposed_method},  result tables for the other data sets and other similarity measures can be found in
the supplementary material.
Before we present the learning results, we will discuss the structure of the label similarity matrices for the data sets ImageNet, CIFAR100, and AWA2.

\paragraph{Label Similarities.}
For a better understanding of the label similarity matrices, we visualized their eigenspectra in Figure~\ref{fig:eig_imagenet_class_dis_matrix}. 
Consider two extreme scenarios:
If a label similarity matrix has rank 1, all classes are exactly the same and there cannot be any discriminatory learning. In contrast,  the full rank case with equal eigenvalues is the standard learning case where all classes are orthogonal  to each other (one-hot-encoding). 
Figure~\ref{fig:eig_imagenet_class_dis_matrix}
shows  exponential eigenvalues decays, which means there are clusters of similar classes. Distinguishing between these clusters of classes is an easier task than distinguishing between classes within one cluster. 

\begin{table}[ht!]
{
    \begin{center}
{
\caption{ImageNet. Top-1 results, averaged over four trials.}
\label{Tab:top_1_ImageNet}
\begin{tabular}{lccccccc}
{} & {} & {} & {} & {} &\multicolumn{3}{c}{$\boldsymbol{\epsilon}$}\\
\cmidrule{6-8}
{} & \textbf{SL} &             \textbf{LS} & \textbf{KD} & \textbf{CN} &          $\boldsymbol{0.9}$ &           $\boldsymbol{0.99}$ &     $\boldsymbol{0.999}$ \\
\midrule
&\multicolumn{7}{c}{DR = 5\%}\\
\cmidrule{2-8}
\ResnetS          &    38.21 &   38.43 & 39.81 & 36.12 & 41.24 &    41.6 &  \textbf{42.21} \\
\ResnextSshort &   45.46 &   45.71 &  46.21 & 44.14 &  46.1 &   46.92 &  \textbf{47.12} \\
\SENet         &   48.29 &   48.57 &   48.44 & 46.21 & 49.8 &   50.04 & \textbf{50.19} \\
\cmidrule{2-8}
&\multicolumn{7}{c}{DR = 10\%}\\
\cmidrule{2-8}
\ResnetS          &   51.95 &   52.25 &  53.64 & 52.17 & 55.39 &   55.62 &  \textbf{55.64} \\
\ResnextSshort &   58.63 &   58.92 &   58.94 & 57.64 & 59.78 &   \textbf{60.07} &  59.92 \\
\SENet        &   60.61 &   60.74 & 60.82 & 60.14 & 60.99 &   61.18 &  \textbf{62.28} \\
\cmidrule{2-8}
&\multicolumn{7}{c}{DR = 20\%}\\
\cmidrule{2-8}
\ResnetS          &   61.87 &    62.11 &   63.17 & 62.41 & 64.41 &   64.42 &   \textbf{64.44} \\
\ResnextSshort &   67.96 &   68.13 & 68.29 & 68.14 &  68.48 &   68.47 &  \textbf{68.57} \\
\SENet         &   67.77 &   67.71 & 67.64 & 67.43 &  68.14 &   \textbf{68.4} &  68.33 \\
\cmidrule{2-8}
&\multicolumn{7}{c}{DR = 100\%}\\
\cmidrule{2-8}
\ResnetS          &   76.25 &    76.4 & 76.38 & 76.1 & 76.71 &   76.75 &  \textbf{76.89} \\
\ResnextSshort &   78.05 &  78.17  & 78.21 & 77.94 & 78.31 &    78.5 &  \textbf{78.64} \\
\SENet         &   79.33 &   79.65 &   79.44 & 79.44 & 80.11 &   80.03 &  \textbf{80.21} \\
\bottomrule
\end{tabular}
}
\end{center}
}
\end{table}

\begin{table}[ht!]
\begin{center}
\caption{ImageNet. Top-5 results, averaged over four trials.}
\label{Tab:top_5_ImageNet}
\begin{tabular}{lccccccc}
{} & {} & {} & {} & {} &\multicolumn{3}{c}{$\boldsymbol{\epsilon}$}\\
\cmidrule{6-8}
{} & \textbf{SL} &             \textbf{LS} &    \textbf{KD} & \textbf{CN} &        $\boldsymbol{0.9}$ &           $\boldsymbol{0.99}$ &     $\boldsymbol{0.999}$ \\
\midrule
&\multicolumn{7}{c}{DR = 5\%}\\
\cmidrule{2-8}
\ResnetS          &  64.04& 64.35& 67.22& 65.12
 & 64.41 & 67.59&  \textbf{67.94}\\
\ResnextSshort           &  70.52& 70.76& 71.84& 70.92 & 70.67 & 72.05&  \textbf{72.18}\\
\SENet            &  73.35& 73.52& 74.54& 73.92
 & 73.56 & 74.65&  \textbf{74.92}\\
\cmidrule{2-8}
&\multicolumn{7}{c}{DR = 10\%}\\
\cmidrule{2-8}
\ResnetS          &  76.86&  77.04& 79.73& 78.14 & 77.1 & \textbf{79.8}&   79.69\\
\ResnextSshort           &  81.52&  81.87& 82.56& 81.92 &	81.67 & 82.66&  \textbf{82.74}\\
\SENet             &  82.53&  \textbf{83.71}& 83.38& 82.76 &	82.94 & 83.6&   83.5\\
\cmidrule{2-8}
&\multicolumn{7}{c}{DR = 20\%}\\
\cmidrule{2-8}
\ResnetS          &  84.41&  84.57& 86.1& 85.36 &	84.91 &  \textbf{86.15}&  86.14\\
\ResnextSshort           &  87.96&  88.11& 88.2& 88.04	& 87.84 & \textbf{88.36}&  \textbf{88.36}\\
\SENet            &  88.16&  88.23& 88.17& 88.17 &	88.11 & \textbf{88.37}&  88.31\\
\cmidrule{2-8}
&\multicolumn{7}{c}{DR = 100\%}\\
\cmidrule{2-8}
\ResnetS          &  92.87&  92.91& 92.94& 92.41 &	92.84 & \textbf{92.95}&  92.93\\
\ResnextSshort           &  93.95&  93.92& 93.96& 93.85 &	93.87 & 94.07&  \textbf{96.15}\\
\SENet            &  94.33&  94.44& 94.84& 94.02 &	94.35 & \textbf{94.93}&  94.79\\
\bottomrule
\end{tabular}
\end{center}
\end{table}

\begin{table}[ht]
    \begin{center}
\caption{\cifar{}. Top-1 results, averaged over four trials.}
\label{Tab:top_1_CIFAR100}
\begin{tabular}{lccccccccc}
{} & {} & {}& {} &{}& {} &{} &\multicolumn{3}{c}{$\boldsymbol{\epsilon}$}\\
\cmidrule{8-10}
{} & \textbf{SL} &            \textbf{LS} &      \textbf{DML}$_{\boldsymbol{1}}$ &  \textbf{DML}$_{\boldsymbol{2}}$ &   \textbf{KD} & \textbf{CN} &          $\boldsymbol{0.9}$ &           $\boldsymbol{0.99}$ &     $\boldsymbol{0.999}$\\
\midrule
&\multicolumn{9}{c}{DR = 5\%}\\
\cmidrule{2-10}
\WRNshort              &    40.2 &  40.31 &  40.16 & 40.47 & 40.94 & 38.14 & 41.36&  41.86 & \textbf{41.92} \\
\DenseNetshort    & 43.34 & 43.5 &  43.89 & 44.14 & 43.76 & 43.42 &  \textbf{44.66}& 45.37&  44.53\\
\cmidrule{2-10}
&\multicolumn{9}{c}{DR = 10\%}\\
\cmidrule{2-10}
\WRNshort               & 60.2 &   60.1 &  60.38 &  60.34 & 60.45 & 60.14  &  60.49 &  60.86&  \textbf{61.19} \\
\DenseNetshort &   60.85&   61.1 &  61.22 &  61.34 & 60.81 & 59.83  &   61.5 & 61.4&  \textbf{61.65} \\
\cmidrule{2-10}
&\multicolumn{9}{c}{DR = 20\%}\\
\cmidrule{2-10}
\WRNshort               &   71.05 &  71.25& 71.61 &  71.65 & 71.53 & 71.37  &  71.64& 71.67 &  \textbf{
71.83} \\
\DenseNetshort &   72.38 & 72.39 &   71.5&  71.34 & 72.24 & 71.54 &  72.71&  72.65 &  \textbf{72.87} \\
\cmidrule{2-10}
&\multicolumn{9}{c}{DR = 100\%}\\
\cmidrule{2-10}
\WRNshort               &   79.52 &  79.84 & 80.32&  80.20 & 80.14 & 78.64    &  81.17 &  81.15 &  \textbf{81.25} \\
\DenseNetshort &   82.85 &  83.01 &  82.91 &  82.57 & 82.67 & 81.14 &  82.96 &  83.11 &   \textbf{83.2} \\
\bottomrule
\end{tabular}
\end{center}
\end{table}

\begin{table}[ht!]
    \begin{center}
    \caption{\cifar{}. Top-5 results, averaged over four trials.}
\label{Tab:top_5_CIFAR100}
    {\begin{tabular}{lccccccccc}
{} & {} & {}& {} &{} & {} &{} &\multicolumn{3}{c}{$\boldsymbol{\epsilon}$}\\
\cmidrule{8-10}
{} & \textbf{SL} &            \textbf{LS} &    \textbf{DML}$_{\boldsymbol{1}}$ &  \textbf{DML}$_{\boldsymbol{2}}$ &     \textbf{KD} & \textbf{CN} &         $\boldsymbol{0.9}$ &           $\boldsymbol{0.99}$ &     $\boldsymbol{0.999}$\\
\midrule
&\multicolumn{9}{c}{DR = 5\%}\\
\cmidrule{2-10}
\WRNshort               &   68.82  &  68.95  &  68.74  &  68.89  & 68.94 &	69.12 &   69.47&  69.39  &  \textbf{69.63}  \\
\DenseNetshort          &   70.61  &  70.85  &   70.7  &  70.92  & 71.14 &	71.27 &   72.1 &  71.13  &  \textbf{72.52}  \\
\cmidrule{2-10}
&\multicolumn{9}{c}{DR = 10\%}\\
\cmidrule{2-10}
\WRNshort               &   83.64  &  83.82  &  84.05  &  84.11 & 83.94 &	83.77 &  83.99 &  84.15  &   \textbf{84.3}  \\
\DenseNetshort          &   84.07  &  84.21  &  84.27  &  84.45 & 84.07 & 84.31 &  84.34 &  \textbf{84.63}  &  84.52  \\
\cmidrule{2-10}
&\multicolumn{9}{c}{DR = 20\%}\\
\cmidrule{2-10}
\WRNshort               &   90.52  &  90.38  &   90.3  &  90.27  & 90.71 & 90.45 &  91.02 &  \textbf{91.19}  &   90.95 \\
\DenseNetshort         &   91.24  &  91.37  &  91.33  &  91.30  & 91.39 & 91.38 &   91.4 &  91.31  &  \textbf{91.47}  \\
\cmidrule{2-10}
&\multicolumn{9}{c}{DR = 100\%}\\
\cmidrule{2-10}
\WRNshort              &   94.04  &  94.23  &  94.44  &  94.42  & 94.52 &	94.52 &   95.29&  95.17  &  \textbf{95.49}  \\
\DenseNetshort          &   95.22  &  95.28  &  95.34  &  95.27  & 95.37 &	95.63 &  95.72 &   95.74 &  \textbf{95.88}  \\
\bottomrule
\end{tabular}
}
\end{center}
\end{table}

\begin{table}[ht!]
    \begin{center}{\caption{AWA2. Top-1 results, averaged over four trials.}
\label{Tab:top_1_AWA2}
\begin{tabular}{lccccccccc}
{} & {} & {}& {} &{} & {} &{} &\multicolumn{3}{c}{$\boldsymbol{\epsilon}$}\\
\cmidrule{8-10}
{} & \textbf{SL} &            \textbf{LS} &       \textbf{DML}$_{\boldsymbol{1}}$ &  \textbf{DML}$_{\boldsymbol{2}}$ &  \textbf{KD} & \textbf{CN} &             $\boldsymbol{0.9}$ &           $\boldsymbol{0.99}$ &     $\boldsymbol{0.999}$\\
\midrule
&\multicolumn{9}{c}{DR = 5\%}\\
\cmidrule{2-10}
\ResnetB         &   23.09  &  27.82 &  39.22 &  37.19 & 41.58       & 24.1  &  45.51 &  45.55  &  \textbf{45.78}   \\
\inceptionresnet &   57.42 &  57.95  &  58.69 &  58.14 & 59.3        & 56.9 &  60.85 &  \textbf{61.07}  &    60.71 \\
\cmidrule{2-10}
&\multicolumn{9}{c}{DR = 10\%}\\
\cmidrule{2-10}
\ResnetB         &   41.86 &  44.98  &  48.92 &   50.5& 44.02       & 43.12 &  47.21 &  51.67  &   \textbf{53.39}  \\
\inceptionresnet &    71.47&  71.86  &  71.82 &  72.37 & 71.49       & 72.01 &  72.61 &  72.97  &   \textbf{73.01}  \\
\cmidrule{2-10}
&\multicolumn{9}{c}{DR = 20\%}\\
\cmidrule{2-10}
\ResnetB         &   77.11 &  78.23  &   78.34&  78.32 & 78.28       & 77.64 &  80.03 &  \textbf{80.07}  &   79.86  \\
\inceptionresnet &   83.64 &  83.92  &  83.87 &  83.76 & 83.83       & 84.12 &  \textbf{84.27} &  84.05  &   \textbf{84.27}  \\
\cmidrule{2-10}
&\multicolumn{9}{c}{DR = 100\%}\\
\cmidrule{2-10}
\ResnetB         &   88.73 &  89.25  &  89.01 &  89.11  & 89.17       & 88.92  &  89.44 &  \textbf{89.64}  &   89.63  \\
\inceptionresnet &   89.69 &  89.94  &  90.05 &  90.22 & 89.94       & 89.29 &  \textbf{90.49} &  90.34  &   90.47  \\
\bottomrule
\end{tabular}
}
\end{center}

\end{table}

\begin{table}[ht!]
    \begin{center}{\caption{AWA2. Top-5 results, averaged over four trials.}
\label{Tab:top_5_AWA2}
\begin{tabular}{lccccccccc}
{} & {} & {}& {} &{}& {} &{} &\multicolumn{3}{c}{$\boldsymbol{\epsilon}$}\\
\cmidrule{8-10}
{} & \textbf{SL} &            \textbf{LS} &    \textbf{DML}$_{\boldsymbol{1}}$ &  \textbf{DML}$_{\boldsymbol{2}}$ &  \textbf{KD} & \textbf{CN} &             $\boldsymbol{0.9}$ &      $\boldsymbol{0.99}$ &         $\boldsymbol{0.999}$ \\
\midrule
&\multicolumn{9}{c}{DR = 5\%}\\
\cmidrule{2-10}
\ResnetB         &   53.31  &  54.19 &  65.14 &    63.12 &  56.19 & 54.17  &76.02  &  76.14  &  \textbf{76.47}  \\
\inceptionresnet &   83.06  &  83.14  &  84.07  &  84.18 & 84.12 &	83.77 & 84.94  &  \textbf{85.24}  &  84.84  \\
\cmidrule{2-10}
&\multicolumn{9}{c}{DR = 10\%}\\
\cmidrule{2-10}
\ResnetB         &   72.59  &  72.43  &  75.07  &  76.14 & 76.61 &	76.34 & 77.04  &  \textbf{80.46}  &  80.11  \\
\inceptionresnet &   91.37  &  91.42  &  91.35  &  91.43 & 91.48	& 91.35 &  \textbf{91.9}  &  91.89  &   91.71 \\
\cmidrule{2-10}
&\multicolumn{9}{c}{DR = 20\%}\\
\cmidrule{2-10}
\ResnetB         &   94.21  &  94.56  &  94.79  &  95.01 &  94.61	& 94.45 & \textbf{95.2}  &  95.07  &  95.12  \\
\inceptionresnet &   96.03  &  96.23  &  96.28  &  96.13 &  96.21 &	95.19 &96.18  &  96.49  &  \textbf{96.57}  \\
\cmidrule{2-10}
&\multicolumn{9}{c}{DR = 100\%}\\
\cmidrule{2-10}
\ResnetB         &   97.85  &  97.92  &   98.1  &  97.95 & 97.43 &	97.32 & 98.11  &  \textbf{98.14}  &   98.1  \\
\inceptionresnet &   98.01  &  98.07  &  98.25  &   98.17& 97.67 &	97.56 & 98.25  &  \textbf{98.41}  &   98.2  \\
\bottomrule
\end{tabular}
}
\end{center}

\end{table}

\paragraph{Classification Performance.}
We measured the  top-$1$ classification accuracy and top-$5$ classification
accuracy after the last epoch. The results are summarized in Table \ref{Tab:top_1_ImageNet} and Table \ref{Tab:top_5_ImageNet} for ImageNet, in Table \ref{Tab:top_1_CIFAR100} and Table \ref{Tab:top_5_CIFAR100} for \cifar, and in Table \ref{Tab:top_1_AWA2} and Table \ref{Tab:top_5_AWA2} for AWA2 (for standard deviations see the supplementary material). All results are averaged over four trials.
It is important to keep in mind that we compare against baseline
results achieved with architectures and hyperparameters tuned for
excellent performance. 
Furthermore, we compare to baseline results from our own experiments,
not to results taken from the literature. We ran each experiment 4 times with same seeds for all algorithms. This allows for a fair comparison. Our averaged results also provide a more reliable estimate of the performance of the systems compared to single trials reported in the original works.





The results show that for all datasets and in  all experimental cases using {LCL} outperformed all baselines, with SeNet with $\DR=10\%$ and top-5 metric being the only exception. The improvement was more pronounced when $\DR < 100\%$. It is quite intuitive that a curriculum is much more important when the training data is limited (i.e., the learning problem is more difficult). Loosely speaking, the importance of a teacher decreases when a student has access to unlimited  information without any computational and/or time budget.

For example, for \ResnetS on ImageNet  {LCL} improved the top-1 accuracy on average by $4$ percentage points (p.p.) over the baseline when $\DR = 5\%$, and  2\,p.p.\ in top-5 accuracy were gained with  $\DR = 100\%$ on ImageNet for the ResNeXt architecture.
The biggest improvements were achieved on the AWA2 dataset.
For \ResnetB and $\DR = 5\%$,  average improvements of  more than 22\,p.p.\ and 23\,p.p.\ could be achieved  in the top-1 and top-5 accuracy, respectively.
As could be expected, the performance gains in the top-$5$ setting were typically smaller than for top-$1$. Still, nothing changed with respect to the ranking of the network architectures. 

Larger values of $\epsilon$ mean slower convergence to the
one-hot-encoding and therefore more emphasis on the curriculum
learning. In most experiments, 
$\epsilon = 0.999$ performed best.
The observation that larger $\epsilon$ values gave better results than small ones
provides additional evidence that the curriculum really
supports the learning process (and that we are not discussing random artifacts).

Under the assumption that the experimental scenarios are statistically independent, we performed a 
{statistical comparisons of the classifiers over
multiple data sets for all pairwise comparisons} following \cite{demvsar2006scc,garcia2008esc}.
Using the Iman and Daveport test, all but one result were statistically significant. If we consider all  ImageNet top-1 accuracy results, our method with $\epsilon=0.999$  ranked best, followed by $\epsilon=0.99$, $\epsilon=0.9$, LS and then SL. This ranking was highly significant (Iman and Daveport test,  $p<0.001$). Similarly, our method with $\epsilon=0.999$ was best for both CIFAR-100 and AWA2 ($p<0.001$).

\section{Conclusions}
\label{sec:Conclusions}
We proposed a novel curriculum learning approach referred to as
label-similarity curriculum learning.
In contrast to previous methods, which change the weighting of training examples,
it is based on adapting the label representation during training.
This adaptation considers the semantic similarity of labels. It implements  the basic idea that at an early stage of learning it is less important to
distinguish between similar classes compared to separating very different classes.

The class similarity can be based on arbitrary \emph{a  priori} knowledge, in particular on additional information not directly encoded in the training data.
For the case where the class labels are natural language words, we 
proposed a way to  automatically define class similarity via a word embedding. We also  considered other similarity measures for  datasets where these similarity measures were available.

We extensively evaluated the approach on five datasets. For each dataset, two to three deep learning architectures proposed in the literature were considered. We looked at simple  label smoothing and, for the two smaller datasets, also at deep mutual learning (DML) as additional baselines. In each case, we considered four different training data set sizes. Each experiment was repeated four times.
The empirical results strongly support our approach. \emph{Label-similarity curriculum learning 
was able to improve the average classification accuracy on the test data compared to standard training in all scenarios.} 
The improvements achieved by our method were more pronounced for smaller training data sets.
When considering only $10\%$ of the AWA2 training data, label-similarity curriculum learning increased 
the Resnet101 top-1 test accuracy  by  more than 22 percentage points on average compared to the standard baseline.

Our curriculum learning also outperformed simple label smoothing and DML in all but a single case.  Our method turned out to be robust with respect to the choice of the single hyperparameter controlling how quickly the learning process converges to minimizing the standard cross-entropy loss. In contrast to related approaches such as knowledge distillation and DML, the additional computational and memory requirements can be neglected. 

The proposed  label-similarity curriculum learning is a general approach, 
which also works for settings where the class similarity is not based on the semantic similarity of natural language words (see supplementary material).
\clearpage
\bibliographystyle{splncs04}
\bibliography{curriculum}

\clearpage
\renewcommand\thefigure{\thesection.\arabic{figure}} 
\renewcommand\thetable{\thesection.\arabic{table}} 
\renewcommand\theequation{\thesection.\arabic{equation}} 

\appendix 
\label{sec:appendix}


 
\section{Standard Deviations for Results}

\begin{table}[ht]
{
    \begin{center}
    \caption{ImageNet. Top-1 results, standard deviations over four trials.}
\label{Tab:top_1_ImageNet_std}
{
\begin{tabular}{lccccccc}
{} & {} & {} & {} & {} &\multicolumn{3}{c}{$\boldsymbol{\epsilon}$}\\
\cmidrule{6-8}
{} & \textbf{SL} &             \textbf{LS} & \textbf{KD} & \textbf{CN}    &       $\boldsymbol{0.9}$ &           $\boldsymbol{0.99}$ &     $\boldsymbol{0.999}$ \\
\midrule
&\multicolumn{7}{c}{DR = 5\%}\\
\cmidrule{2-8}
\ResnetS          &   0.6 &   0.47  & 0.08 &	0.17 & 0.89 &    0.62 &  \textbf{0.78} \\
\ResnextSshort &   0.46 &  0.53 &  0.2	& 0.2 &  0.15 &  0.07 &  \textbf{0.11} \\
\SENet         &   0.34 &   0.41 &  0.15 & 0.11 & 0.21 &  0.09 & \textbf{0.25} \\
\cmidrule{2-8}
&\multicolumn{7}{c}{DR = 10\%}\\
\cmidrule{2-8}
\ResnetS          &   1.31 &   0.64 &  0.21 & 0.42 & 0.16 &   0.27 &  \textbf{0.14} \\
\ResnextSshort &   0.39 &   0.27 &  0.13 & 0.28 &	0.24 &  \textbf{0.08} &  0.16 \\
\SENet        &   0.02 &   0.3 &  0.23 & 0.28
 & 0.15 &   0.21 &  \textbf{0.23} \\
\cmidrule{2-8}
&\multicolumn{7}{c}{DR = 20\%}\\
\cmidrule{2-8}
\ResnetS          &   0.1 &    0.34  &  0.34 &	0.3 & 0.08  &   0.18 &   \textbf{0.1} \\
\ResnextSshort &  0.23 &   0.37 & 0.06	& 0.3 &  0.12 &   0.07 &  \textbf{0.14} \\
\SENet         &   0.15 &   0.23 &  0.19	& 0.28 & 0.13 &   \textbf{0.14} &  0.02 \\
\cmidrule{2-8}
&\multicolumn{7}{c}{DR = 100\%}\\
\cmidrule{2-8}
\ResnetS          &   0.14 &    0.17 & 0.22	& 0.25 &  0.12 &   0.14 &  \textbf{0.11} \\
\ResnextSshort &   0.05 &  0.23  & 0.42 & 0.25
 & 0.04 &    0.06 &  \textbf{0.08} \\
\SENet        &   0.12 &   0.24  & 0.19 & 0.31 & 0.14 &  0.11 &  \textbf{0.14} \\
\bottomrule
\end{tabular}
}
\end{center}
}
\end{table}

\begin{table}[ht]
{
    \begin{center}
    \caption{ImageNet. Top-5 results, averaged over four trials.}
\label{Tab:top_5_ImageNet_std}
{
\begin{tabular}{lccccccc}

{} & {} & {}  & {} & {} &\multicolumn{3}{c}{$\boldsymbol{\epsilon}$}\\
\cmidrule{6-8}
{} & \textbf{SL} &             \textbf{LS} & \textbf{KD} & \textbf{CN} &           $\boldsymbol{0.9}$ &           $\boldsymbol{0.99}$ &     $\boldsymbol{0.999}$ \\
\midrule
&\multicolumn{7}{c}{DR = 5\%}\\
\cmidrule{2-8}
\ResnetS          &  0.81 & 0.92 & 0.49	& 0.31 & 1.06 & 0.66 &  \textbf{0.73}\\
\ResnextSshort           & 0.4 & 0.72 & 0.41	& 0.29
& 0.18 & 0.1 &  \textbf{0.18}\\
\SENet             &  0.56 & 0.41& 0.29 &	0.31 & 0.13 & 0.22 &  \textbf{0.26}\\
\cmidrule{2-8}
&\multicolumn{7}{c}{DR = 10\%}\\
\cmidrule{2-8}
\ResnetS          &  1.02 &  0.86 & 0.62 &	0.34 &0.16 & \textbf{0.17 }&  0.07 \\
\ResnextSshort           & 0.2 &  0.21 & 0.55 &	0.79
 & 0.05 & 0.14 &  \textbf{0.11}\\
\SENet           &  0.1 &  \textbf{0.37}& 0.81 & 0.48 & 0.39 & 0.2 &   0.32 \\
\cmidrule{2-8}
&\multicolumn{7}{c}{DR = 20\%}\\
\cmidrule{2-8}
\ResnetS          &  0.04 &  0.42 & 0.58 &	0.31 & 0.08 &  \textbf{0.08}&  0.06 \\
\ResnextSshort           &  0.16 &  0.22 & 0.5	& 0.78
 & 0.13 &  \textbf{0.1}&  \textbf{0.08}\\
\SENet            &  0.05 &  0.17 & 0.32 &	0.42 & 0.15 & \textbf{0.1}&  0.11\\
\cmidrule{2-8}
&\multicolumn{7}{c}{DR = 100\%}\\
\cmidrule{2-8}
\ResnetS          &  0.09 &  0.1 & 0.18 &	0.39
& 0.07 & \textbf{0.09}&  0.06\\
\ResnextSshort           &  0.04 &  0.1 & 0.57	& 0.37
& 0.05 & 0.03 &  \textbf{0.04}\\
\SENet            &  0.27 &  0.28 & 0.29 &	0.19 & 0.15 & \textbf{0.18}&  0.22\\
\bottomrule
\end{tabular}
}
\end{center}
}
\end{table}

\begin{table}[ht]
    \begin{center}
    \caption{\cifar{}. Top-1 results, standard deviations over four trials.}
\label{Tab:top_1_CIFAR100_std}
{
\begin{tabular}{lccccccccc}

{} & {} & {}& {} &{}& {} &{} &\multicolumn{3}{c}{$\boldsymbol{\epsilon}$}\\
\cmidrule{8-10}
{} & \textbf{SL} &            \textbf{LS} &     \textbf{DML}$_{\boldsymbol{1}}$ &  \textbf{DML}$_{\boldsymbol{2}}$&   \textbf{KD} & \textbf{CN} &        $\boldsymbol{0.9}$ &           $\boldsymbol{0.99}$ &     $\boldsymbol{0.999}$\\
\midrule
&\multicolumn{9}{c}{DR = 5\%}\\
\cmidrule{2-10}
\WRNshort              &    0.67 &  0.78 &  0.9 & 0.63& 0.58 & 0.24 & 0.41&  0.81 & \textbf{0.6} \\
\DenseNetshort    & 0.42 & 0.61 &  0.45 & 0.35 & 0.19 & 0.13 & \textbf{0.57}& 0.51 &  0.59 \\
\cmidrule{2-10}
&\multicolumn{9}{c}{DR = 10\%}\\
\cmidrule{2-10}
\WRNshort               & 0.1 &   0.6 &  0.29 &  0.36 & 0.24	& 0.28 & 0.17 &  0.13 &  \textbf{0.31} \\
\DenseNetshort &   0.44 &   0.53 &  0.18 &  0.21 & 0.28 & 0.36 &  0.28 & 0.29 &  \textbf{0.17} \\
\cmidrule{2-10}
&\multicolumn{9}{c}{DR = 20\%}\\
\cmidrule{2-10}
\WRNshort               &   0.67 &  0.21 & 0.4 &  0.26 &  0.14& 0.35 & 0.3 & 0.2 &  \textbf{0.23} \\
\DenseNetshort &   0.23  & 0.41 &   0.15 &  0.21 & 0.41 & 0.21 &  0.18 & 0.18 &  \textbf{0.17} \\
\cmidrule{2-10}
&\multicolumn{9}{c}{DR = 100\%}\\
\cmidrule{2-10}
\WRNshort               &   0.21  &  0.23 & 0.14 &  0.28 & 0.51& 0.19 & 0.12 &  0.08 &  \textbf{0.11} \\
\DenseNetshort &   0.27 &  0.65 &  0.21 &  0.32 & 0.37& 0.34 & 0.17 &  0.11 &   \textbf{0.07} \\
\bottomrule
\end{tabular}
}
\end{center}
\end{table}

\begin{table}[ht]
    \begin{center}
    \caption{\cifar{}. Top-5 results, standard deviations over four trials.}
\label{Tab:top_5_CIFAR100_std}
{
\begin{tabular}{lccccccccc}

{} & {} & {}& {} &{}& {} &{} &\multicolumn{3}{c}{$\boldsymbol{\epsilon}$}\\
\cmidrule{8-10}
{} & \textbf{SL} &            \textbf{LS} &     \textbf{DML}$_{\boldsymbol{1}}$ &  \textbf{DML}$_{\boldsymbol{2}}$ &      
\textbf{KD} & \textbf{CN} &  $\boldsymbol{0.9}$ &           $\boldsymbol{0.99}$ &     $\boldsymbol{0.999}$\\
\midrule
&\multicolumn{9}{c}{DR = 5\%}\\
\cmidrule{2-10}
\WRNshort               &  0.88  &  0.57  &  0.63  &  0.41  & 0.56	& 0.14 &  0.3&  0.29  &  \textbf{0.32}  \\
\DenseNetshort          &   0.64  &  0.54  &   0.27  &  0.33  & 0.41& 0.14 &  1.02 &  0.57  &  \textbf{0.44}  \\
\cmidrule{2-10}
&\multicolumn{9}{c}{DR = 10\%}\\
\cmidrule{2-10}
\WRNshort               &   0.32  &  0.45  &  0.23  &  0.24  &  0.21 & 0.24 &0.18 &  0.15  &   \textbf{0.11}  \\
\DenseNetshort          &   0.29  &  0.32  &  0.25  &  0.37  & 0.25 & 0.39 & 0.18 &  \textbf{0.11}  &  0.17  \\
\cmidrule{2-10}
&\multicolumn{9}{c}{DR = 20\%}\\
\cmidrule{2-10}
\WRNshort               &   0.31  &  0.71  &   0.21  &  0.27  & 0.24 & 0.42 & 0.25  &  \textbf{0.25}  &   0.2 \\
\DenseNetshort          &   0.12  &  0.54  &  0.11  &  0.17 &  0.32 & 0.27 &  0.19 &  0.36  &  \textbf{0.14}  \\
\cmidrule{2-10}
&\multicolumn{9}{c}{DR = 100\%}\\
\cmidrule{2-10}
\WRNshort               &   0.18  &  0.55  &  0.24  &  0.24  &  0.39 & 0.24 & 0.1 &  0.14  &  \textbf{0.06}  \\
\DenseNetshort          &   0.22  &  0.38  &  0.14  &  0.24  & 0.36 & 0.29 & 0.08 &   0.1 &  \textbf{0.04}  \\
\bottomrule
\end{tabular}
}
\end{center}
\end{table}

\begin{table}[ht]
    \begin{center}
    \caption{AWA2. Top-1 results, standard deviations over four trials.}
\label{Tab:top_1_AWA2_std}
{
\begin{tabular}{lccccccccc}

{} & {} & {}& {} &{}& {} &{} &\multicolumn{3}{c}{$\boldsymbol{\epsilon}$}\\
\cmidrule{8-10}
{} & \textbf{SL} &            \textbf{LS} &    \textbf{DML}$_{\boldsymbol{1}}$ &  \textbf{DML}$_{\boldsymbol{2}}$ &
\textbf{KD} & \textbf{CN} & $\boldsymbol{0.9}$ &           $\boldsymbol{0.99}$ &     $\boldsymbol{0.999}$\\
\midrule
&\multicolumn{9}{c}{DR = 5\%}\\
\cmidrule{2-10}
\ResnetB         &   9.15  &  6.72 &  4.94 &  5.42 &  0.1 & 0.25 & 1.74  &  1.99  &  \textbf{1.14}   \\
\inceptionresnet &   1.17 &  1.43  &  1.21 &  0.94 & 0.21 & 0.079 & 0.71 &  \textbf{0.93}  &    0.6 \\
\cmidrule{2-10}
&\multicolumn{9}{c}{DR = 10\%}\\
\cmidrule{2-10}
\ResnetB         &   8.21 &  5.27  &  3.5 &   2.85 &  0.16 & 0.1 & 4.78  &  2.85  &   \textbf{2.91}  \\
\inceptionresnet &    0.8&  0.95  & 0.45 &  0.63 & 0.11	& 0.11 & 0.34 &  0.58  &   \textbf{0.25}  \\
\cmidrule{2-10}
&\multicolumn{9}{c}{DR = 20\%}\\
\cmidrule{2-10}
\ResnetB         &   1.44 &  1.45  &   0.4 &  0.83 &  0.23 & 0.27 & 0.88  &  \textbf{1.02}  &   0.73 \\
\inceptionresnet &   0.19 &  0.34  &  0.21 &  0.24 & 0.26 & 0.41 &  \textbf{0.18} &  0.16  &   \textbf{0.22}  \\
\cmidrule{2-10}
&\multicolumn{9}{c}{DR = 100\%}\\
\cmidrule{2-10}
\ResnetB         &   0.68 &  0.54  &  0.42 &  0.45 &  0.32 & 0.12 & 0.37 &  \textbf{0.38}  &   0.24  \\
\inceptionresnet &   0.32 &  0.21  &  0.16 &  0.09 &  0.32 & 0.46 & \textbf{0.11} &  0.15  &   0.11  \\
\bottomrule
\end{tabular}
}
\end{center}

\end{table}

\begin{table}[ht]
\caption{AWA2. Top-5 results, standard deviations over four trials.}
\label{Tab:top_5_AWA2_std}
    \begin{center}
{
\begin{tabular}{lccccccccc}

{} & {} & {}& {} &{}& {} &{} &\multicolumn{3}{c}{$\boldsymbol{\epsilon}$}\\
\cmidrule{8-10}
{} & \textbf{SL} &           \textbf{ LS }&     \textbf{DML}$_{\boldsymbol{1}}$ &  \textbf{DML}$_{\boldsymbol{2}}$ &  
\textbf{KD} & \textbf{CN} &$0.9$ &      $0.99$ &         $0.999$ \\
\midrule
&\multicolumn{9}{c}{DR = 5\%}\\
\cmidrule{2-10}
\ResnetB         &  11.73  &  9.82 &  4.79  &    4.12 &  0.29 & 0.34 & 1.77   &  1.97  &  \textbf{0.79}  \\
\inceptionresnet &   0.64  &  0.65  &  0.54  & 0.13  & 0.37	& 0.16 & 0.61  &  \textbf{0.81}  &  0.21  \\
\cmidrule{2-10}
&\multicolumn{9}{c}{DR = 10\%}\\
\cmidrule{2-8}
\ResnetB         &   7.84  &  0.45  &  3.02  &  2.19 & 0.34	& 0.3 &  4.72  &  \textbf{2.01}  &  2.94  \\
\inceptionresnet &   0.22  &  0.23  &  0.38  &  0.44 & 0.25	& 0.31 &   \textbf{0.23}  &  0.27  &   0.3 \\
\cmidrule{2-10}
&\multicolumn{9}{c}{DR = 20\%}\\
\cmidrule{2-10}
\ResnetB         &   0.37  &  0.32  &  0.51  &  0.32 &   0.14 & 0.22 & \textbf{0.28}  &  0.74  &  0.23  \\
\inceptionresnet &   0.07  &  0.27  &  0.07  &  0.23 & 0.23	& 0.33 & 0.14  &  0.22  &  \textbf{0.16}  \\
\cmidrule{2-10}
&\multicolumn{9}{c}{DR = 100\%}\\
\cmidrule{2-10}
\ResnetB         &   0.08  &  0.23  &   0.23  &  0.34 &  0.35 & 0.25 & 0.02  &  \textbf{0.18}  &  0.05  \\
\inceptionresnet &   0.12  &  0.29  &  0.11  &  0.2 &  0.24 & 0.3 & 0.06  &  \textbf{0.37}  &   0.17  \\
\bottomrule
\end{tabular}
}
\end{center}
\end{table}

\clearpage
\section{Similarity Measures}
Details on the semantic similarity are given in subsection \ref{subsec:semantic_similarity}. For computing visual similarity, we used attribute vectors provided with the images. For example, an image of an animal in the AWA2 dataset may have a $d$-dimensional attribute vector describing its color, if it has stripes, if it is a water animal, and if it eats fish. Visual similarity is then defined as the cosine similarity between two attribute vectors. Hierarchical similarity for the NABirds data was calculated using simrank~\cite{jeh2002simrank}.

\clearpage
\section{Extra AWA-2 Results}\label{app:awa}

\begin{table}[ht!]
    \begin{center}{\caption{AWA2. Top-1 results, averaged over four trials. We used our method both with the proposed  semantic similarity of natural language labels as well as with a visual similarity measure.}
\label{Tab:top_1_AWA2_extra}
\begin{tabular}{lcccccccccccc}
{} & {} & {}& {} &{} & {} &{} &\multicolumn{3}{c}{$\boldsymbol{\epsilon}$} & \multicolumn{3}{c}{$\boldsymbol{\epsilon}$ visual similarity} \\
\cmidrule{8-13}
{} & \textbf{SL} &            \textbf{LS} &       \textbf{DML}$_{\boldsymbol{1}}$ &  \textbf{DML}$_{\boldsymbol{2}}$ &  \textbf{KD} & \textbf{CN} &             $\boldsymbol{0.9}$ &           $\boldsymbol{0.99}$ &     $\boldsymbol{0.999}$ &             $\boldsymbol{0.9}$ &           $\boldsymbol{0.99}$ &     $\boldsymbol{0.999}$\\
\midrule
&\multicolumn{12}{c}{DR = 5\%}\\
\cmidrule{2-13}
\ResnetB         &   23.09  &  27.82 &  39.22 &  37.19 & 41.58       & 24.1  &  45.51 &  45.55  & {45.78} & 45.74 & \textbf{45.85} & 46.13   \\
\inceptionresnet &   57.42 &  57.95  &  58.69 &  58.14 & 59.3        & 56.9 &  60.85 &  {61.07}  &    60.71 & 61.14& \textbf{61.39} & 60.71 \\
\cmidrule{2-13}
&\multicolumn{12}{c}{DR = 10\%}\\
\cmidrule{2-13}
\ResnetB         &   41.86 &  44.98  &  48.92 &   50.5& 44.02       & 43.12 &  47.21 &  51.67  &   {53.39} & 47.34 &	51.85 & \textbf{53.67}  \\
\inceptionresnet &    71.47&  71.86  &  71.82 &  72.37 & 71.49       & 72.01 &  72.61 &  72.97  &   {73.01} & 72.82 & 73.2 & \textbf{73.34}  \\
\cmidrule{2-13}
&\multicolumn{12}{c}{DR = 20\%}\\
\cmidrule{2-13}
\ResnetB         &   77.11 &  78.23  &   78.34&  78.32 & 78.28       & 77.64 &  80.03 &  {80.07}  &   79.86 & 80.09 &	\textbf{80.8} &	79.94 \\
\inceptionresnet &   83.64 &  83.92  &  83.87 &  83.76 & 83.83       & 84.12 &  {84.27} &  84.05  &   {84.27} & \textbf{85.02}	& 84.47 & 	84.65 \\
\cmidrule{2-13}
&\multicolumn{12}{c}{DR = 100\%}\\
\cmidrule{2-13}
\ResnetB         &   88.73 &  89.25  &  89.01 &  89.11  & 89.17       & 88.92  &  89.44 & {89.64}  &   89.63 & 90.15 &	90.11	& \textbf{90.19} \\
\inceptionresnet &   89.69 &  89.94  &  90.05 &  90.22 & 89.94       & 89.29 &  \textbf{90.49} &  90.34  &   90.47 & 90.27 &	90.33 &	90.46 \\
\bottomrule
\end{tabular}
}
\end{center}

\end{table}

\begin{table}[ht!]
    \begin{center}{\caption{AWA2. Top-5 results, averaged over four trials. We used our method both with the proposed  semantic similarity of natural language labels as well as with a visual similarity measure.}
\label{Tab:top_5_AWA2_extra}
\begin{tabular}{lcccccccccccc}
{} & {} & {}& {} &{} & {} &{} &\multicolumn{3}{c}{$\boldsymbol{\epsilon}$} & \multicolumn{3}{c}{$\boldsymbol{\epsilon}$ visual similarity} \\
\cmidrule{8-13}
{} & \textbf{SL} &            \textbf{LS} &       \textbf{DML}$_{\boldsymbol{1}}$ &  \textbf{DML}$_{\boldsymbol{2}}$ &  \textbf{KD} & \textbf{CN} &             $\boldsymbol{0.9}$ &           $\boldsymbol{0.99}$ &     $\boldsymbol{0.999}$ &             $\boldsymbol{0.9}$ &           $\boldsymbol{0.99}$ &     $\boldsymbol{0.999}$\\
\midrule
&\multicolumn{12}{c}{DR = 5\%}\\
\cmidrule{2-13}
\ResnetB         &   53.31  &  54.19 &  65.14 &    63.12 &  56.19 & 54.17  &76.02  &  76.14  &  \textbf{76.47} & 76.17 &	76.23 &	76.41 \\
\inceptionresnet &   83.06  &  83.14  &  84.07  &  84.18 & 84.12 &	83.77 & 84.94  &  \textbf{85.24}  &  84.84  & 85.02 &	85.17 &	84.78\\
\cmidrule{2-13}
&\multicolumn{12}{c}{DR = 10\%}\\
\cmidrule{2-13}
\ResnetB         &   72.59  &  72.43  &  75.07  &  76.14 & 76.61 &	76.34 & 77.04  &  {80.46}  &  80.11  & 79.54 &	\textbf{80.78}	& 80.56\\
\inceptionresnet &   91.37  &  91.42  &  91.35  &  91.43 & 91.48	& 91.35 &  {91.9}  &  91.89  &   91.71 & 91.65 &	\textbf{91.92}	& 91.53 \\
\cmidrule{2-13}
&\multicolumn{12}{c}{DR = 20\%}\\
\cmidrule{2-13}
\ResnetB         &   94.21  &  94.56  &  94.79  &  95.01 &  94.61	& 94.45 & {95.2}  &  95.07  &  95.12 & \textbf{95.23} &	95.02 &	95.08 \\
\inceptionresnet &   96.03  &  96.23  &  96.28  &  96.13 &  96.21 &	95.19 &96.18  &  96.49  &  {96.57}  & 96.31 &	\textbf{96.67}& 96.43\\
\cmidrule{2-13}
&\multicolumn{12}{c}{DR = 100\%}\\
\cmidrule{2-13}
\ResnetB         &   97.85  &  97.92  &   98.1  &  97.95 & 97.43 &	97.32 & 98.11  &  {98.14}  &   98.1  & 98.43 &	\textbf{98.51} &	98.17 \\
\inceptionresnet &   98.01  &  98.07  &  98.25  &   98.17& 97.67 &	97.56 & 98.25  &  {98.41}  &   98.2 & \textbf{98.64} &	98.44& 98.52 \\
\bottomrule
\end{tabular}
}
\end{center}

\end{table}

\begin{table}[ht]
    \begin{center}
    \caption{AWA2. Top-1 results, standard deviations over four trials. We used our method both with the proposed  semantic similarity of natural language labels as well as with a visual similarity measure.}
\label{Tab:top_1_AWA2_std_extra}
{
\begin{tabular}{lcccccccccccc}

{} & {} & {}& {} &{}& {} &{} &\multicolumn{3}{c}{$\boldsymbol{\epsilon}$} &\multicolumn{3}{c}{$\boldsymbol{\epsilon}$ visual similarity}\\
\cmidrule{8-13}
{} & \textbf{SL} &            \textbf{LS} &    \textbf{DML}$_{\boldsymbol{1}}$ &  \textbf{DML}$_{\boldsymbol{2}}$ &
\textbf{KD} & \textbf{CN} & $\boldsymbol{0.9}$ &           $\boldsymbol{0.99}$ &     $\boldsymbol{0.999}$ & $\boldsymbol{0.9}$ &           $\boldsymbol{0.99}$ &     $\boldsymbol{0.999}$\\
\midrule
&\multicolumn{12}{c}{DR = 5\%}\\
\cmidrule{2-13}
\ResnetB         &   9.15  &  6.72 &  4.94 &  5.42 &  0.1 & 0.25 & 1.74  &  1.99  &  {1.14}  & 0.22	& \textbf{0.18} &	0.11
 \\
\inceptionresnet &   1.17 &  1.43  &  1.21 &  0.94 & 0.21 & 0.079 & 0.71 &  \textbf{0.93}  &    0.6 & 0.24	& \textbf{0.12}	& 0.17
\\
\cmidrule{2-13}
&\multicolumn{12}{c}{DR = 10\%}\\
\cmidrule{2-13}
\ResnetB         &   8.21 &  5.27  &  3.5 &   2.85 &  0.16 & 0.1 & 4.78  &  2.85  &  {2.91} & 0.23& 0.17	& \textbf{0.22}
 \\
\inceptionresnet &    0.8&  0.95  & 0.45 &  0.63 & 0.11	& 0.11 & 0.34 &  0.58  &   {0.25} & 0.19 &	0.15 &	\textbf{0.21}
 \\
\cmidrule{2-13}
&\multicolumn{12}{c}{DR = 20\%}\\
\cmidrule{2-13}
\ResnetB         &   1.44 &  1.45  &   0.4 &  0.83 &  0.23 & 0.27 & 0.88  &  {1.02}  &   0.73 & 0.12	& \textbf{0.18} &	0.19 \\
\inceptionresnet &   0.19 &  0.34  &  0.21 &  0.24 & 0.26 & 0.41 &  {0.18} &  0.16  &   {0.22}  & \textbf{0.15} &	0.23 & 0.17
\\
\cmidrule{2-13}
&\multicolumn{12}{c}{DR = 100\%}\\
\cmidrule{2-13}
\ResnetB         &   0.68 &  0.54  &  0.42 &  0.45 &  0.32 & 0.12 & 0.37 &  {0.38}  &   0.24  & 0.18	& 0.19 &	\textbf{0.16}
 \\
\inceptionresnet &   0.32 &  0.21  &  0.16 &  0.09 &  0.32 & 0.46 & \textbf{0.11} &  0.15  &   0.11  & 0.19	& 0.15 &	0.11
\\
\bottomrule
\end{tabular}
}
\end{center}
\end{table}

\begin{table}[ht]
\caption{AWA2. Top-5 results, standard deviations over four trials. We used our method both with the proposed  semantic similarity of natural language labels as well as with a visual similarity measure.}
\label{Tab:top_5_AWA2_std_extra}
    \begin{center}
{
\begin{tabular}{lcccccccccccc}

{} & {} & {}& {} &{}& {} &{} &\multicolumn{3}{c}{$\boldsymbol{\epsilon}$} &\multicolumn{3}{c}{$\boldsymbol{\epsilon}$ visual similarity}\\
\cmidrule{8-13}
{} & \textbf{SL} &            \textbf{LS} &    \textbf{DML}$_{\boldsymbol{1}}$ &  \textbf{DML}$_{\boldsymbol{2}}$ &
\textbf{KD} & \textbf{CN} & $\boldsymbol{0.9}$ &           $\boldsymbol{0.99}$ &     $\boldsymbol{0.999}$ & $\boldsymbol{0.9}$ &           $\boldsymbol{0.99}$ &     $\boldsymbol{0.999}$\\
\midrule
&\multicolumn{12}{c}{DR = 5\%}\\
\cmidrule{2-13}
\ResnetB         &  11.73  &  9.82 &  4.79  &    4.12 &  0.29 & 0.34 & 1.77   &  1.97  &  \textbf{0.79} & 0.15 &	0.29	& 0.19  \\
\inceptionresnet &   0.64  &  0.65  &  0.54  & 0.13  & 0.37	& 0.16 & 0.61  &  \textbf{0.81}  &  0.21 & 0.28	& 0.52 &	0.11  \\
\cmidrule{2-13}
&\multicolumn{12}{c}{DR = 10\%}\\
\cmidrule{2-13}
\ResnetB         &   7.84  &  0.45  &  3.02  &  2.19 & 0.34	& 0.3 &  4.72  &  {2.01}  &  2.94 & 0.29	& \textbf{0.16}	& 0.12 \\
\inceptionresnet &   0.22  &  0.23  &  0.38  &  0.44 & 0.25	& 0.31 &   {0.23}  &  0.27  &   0.3 & 0.25& \textbf{0.23} &	0.2\\
\cmidrule{2-13}
&\multicolumn{12}{c}{DR = 20\%}\\
\cmidrule{2-13}
\ResnetB         &   0.37  &  0.32  &  0.51  &  0.32 &   0.14 & 0.22 & {0.28}  &  0.74  &  0.23  & \textbf{0.37}	& 0.12 & 	0.17\\
\inceptionresnet &   0.07  &  0.27  &  0.07  &  0.23 & 0.23	& 0.33 & 0.14  &  0.22  &  {0.16} & 0.24	& \textbf{0.23} & 	0.38 \\
\cmidrule{2-13}
&\multicolumn{12}{c}{DR = 100\%}\\
\cmidrule{2-13}
\ResnetB         &   0.08  &  0.23  &   0.23  &  0.34 &  0.35 & 0.25 & 0.02  &  {0.18}  &  0.05 & 0.24 & \textbf{0.27} &	0.28 \\
\inceptionresnet &   0.12  &  0.29  &  0.11  &  0.2 &  0.24 & 0.3 & 0.06  &  {0.37}  &   0.17 & \textbf{0.12}	& 0.18 &	0.17 \\
\bottomrule
\end{tabular}
}
\end{center}
\end{table}

\clearpage

\section{NABirds Results}\label{app:nabirds}
\begin{table}[ht!]
    \begin{center}{\caption{NABirds. Top-1 results, averaged over four trials. The NABirds data set contains images of North American birds. We used our method both with the proposed  semantic similarity of natural language labels as well as with a similarity measure based on the biological hierarchy of the depicted animals.}
\label{Tab:top_1_nabirds}
\begin{tabular}{lcccccccccccc}
{} & {} & {}& {} &{} & {} &{} &\multicolumn{3}{c}{$\boldsymbol{\epsilon}$} & \multicolumn{3}{c}{$\boldsymbol{\epsilon}$ biological hierarchy} \\
\cmidrule{8-13}
{} & \textbf{SL} &            \textbf{LS} &       \textbf{DML}$_{\boldsymbol{1}}$ &  \textbf{DML}$_{\boldsymbol{2}}$ &  \textbf{KD} & \textbf{CN} &             $\boldsymbol{0.9}$ &           $\boldsymbol{0.99}$ &     $\boldsymbol{0.999}$ &             $\boldsymbol{0.9}$ &           $\boldsymbol{0.99}$ &     $\boldsymbol{0.999}$\\
\midrule
\cmidrule{2-13}
&\multicolumn{12}{c}{DR = 100\%}\\
\cmidrule{2-13}
\ResnetVS         & 52.61       & 52.72       & 52.84         & 53.14         & 53.19       & 52.78       & 55.05         & \textbf{55.83}          & 55.53           & 54.11         & \textbf{55.83 }        & 55.16   \\
\ResnetS & 58.81       & 59.19       & 59.17         & 58.76         & 59.64       & 59.04       & 61.46         & 60.16          & 61.17           & 62.68         & 62.19         & \textbf{63.29} \\
\bottomrule
\end{tabular}
}
\end{center}

\end{table}

\begin{table}[ht!]
    \begin{center}{\caption{NABirds. Top-5 results, averaged over four trials. We used our method both with the proposed  semantic similarity of natural language labels as well as with a similarity measure based on the biological hierarchy of the depicted animals.}
\label{Tab:top_5_nabirds}
\begin{tabular}{lcccccccccccc}
{} & {} & {}& {} &{} & {} &{} &\multicolumn{3}{c}{$\boldsymbol{\epsilon}$} & \multicolumn{3}{c}{$\boldsymbol{\epsilon}$ biological hierarchy} \\
\cmidrule{8-13}
{} & \textbf{SL} &            \textbf{LS} &       \textbf{DML}$_{\boldsymbol{1}}$ &  \textbf{DML}$_{\boldsymbol{2}}$ &  \textbf{KD} & \textbf{CN} &             $\boldsymbol{0.9}$ &           $\boldsymbol{0.99}$ &     $\boldsymbol{0.999}$ &             $\boldsymbol{0.9}$ &           $\boldsymbol{0.99}$ &     $\boldsymbol{0.999}$\\
\midrule
\cmidrule{2-13}
&\multicolumn{12}{c}{DR = 100\%}\\
\cmidrule{2-13}
\ResnetVS         & 76.42  &	77.14  &	76.98  &	77.42  &	77.84  &	77.19  &	78.64  &	79.12  &	\textbf{79.45}  &	79.11  &	79.21  &	79.14
   \\
\ResnetS & 82.78 &	83.01 &	83.41&	83.38 &	83.67 &	82.65 &	83.25 &	83.79 &\textbf{	84.29} &	83.67 &	83.31 &	83.75
 \\
\bottomrule
\end{tabular}
}
\end{center}

\end{table}

\begin{table}[ht!]
    \begin{center}{\caption{NABirds. Top-1 results, standard deviations over four trials. We used our method both with the proposed  semantic similarity of natural language labels as well as with a similarity measure based on the biological hierarchy of the depicted animals.}
\label{Tab:top_1_nabirds_std}
\begin{tabular}{lcccccccccccc}
{} & {} & {}& {} &{} & {} &{} &\multicolumn{3}{c}{$\boldsymbol{\epsilon}$} & \multicolumn{3}{c}{$\boldsymbol{\epsilon}$ biological hierarchy} \\
\cmidrule{8-13}
{} & \textbf{SL} &            \textbf{LS} &       \textbf{DML}$_{\boldsymbol{1}}$ &  \textbf{DML}$_{\boldsymbol{2}}$ &  \textbf{KD} & \textbf{CN} &             $\boldsymbol{0.9}$ &           $\boldsymbol{0.99}$ &     $\boldsymbol{0.999}$ &             $\boldsymbol{0.9}$ &           $\boldsymbol{0.99}$ &     $\boldsymbol{0.999}$\\
\midrule
\cmidrule{2-13}
&\multicolumn{12}{c}{DR = 100\%}\\
\cmidrule{2-13}
\ResnetVS         & 0.14  &	0.61  &	0.21  &	0.72	  &0.25  &	0.29  &	0.19  &	\textbf{0.22}  &	0.19  &	0.1	  &\textbf{0.12}  &	0.06
   \\
\ResnetS & 0.32  &	0.23  &	0.19  &	0.45  &	0.33  &	0.23  &	0.24  &	0.21	  &0.17	  &0.11  &	0.09  &	\textbf{0.08}
\\
\bottomrule
\end{tabular}
}
\end{center}

\end{table}

\begin{table}[ht!]
    \begin{center}{\caption{NABirds. Top-5 results, standard deviations over four trials. We used our method both with the proposed  semantic similarity of natural language labels as well as with a similarity measure based on the biological hierarchy of the depicted animals.}
\label{Tab:top_5_nabirds_std}
\begin{tabular}{lcccccccccccc}
{} & {} & {}& {} &{} & {} &{} &\multicolumn{3}{c}{$\boldsymbol{\epsilon}$} & \multicolumn{3}{c}{$\boldsymbol{\epsilon}$ biological hierarchy} \\
\cmidrule{8-13}
{} & \textbf{SL} &            \textbf{LS} &       \textbf{DML}$_{\boldsymbol{1}}$ &  \textbf{DML}$_{\boldsymbol{2}}$ &  \textbf{KD} & \textbf{CN} &             $\boldsymbol{0.9}$ &           $\boldsymbol{0.99}$ &     $\boldsymbol{0.999}$ &             $\boldsymbol{0.9}$ &           $\boldsymbol{0.99}$ &     $\boldsymbol{0.999}$\\
\midrule
\cmidrule{2-13}
&\multicolumn{12}{c}{DR = 100\%}\\
\cmidrule{2-13}
\ResnetVS         & 0.45	 &0.37 &	0.71 &	0.56 &	0.61 &	0.76 &	0.28 &	0.52 &	\textbf{0.22} &	0.17 &	0.34 &	0.16

   \\
\ResnetS & 0.39 &	0.42 &	0.23 &	0.73 &	0.43 &	0.39 &	0.28 &	0.31 &\textbf{	0.27} &	0.32 &	0.23 &	0.22

\\
\bottomrule
\end{tabular}
}
\end{center}
\end{table}

\clearpage 

\section{CUB-200-2011 Results}\label{app:cub}
\begin{table}[ht!]
    \begin{center}{\caption{CUB-200-2011. Top-1 results, averaged over four trials. We used our method both with the proposed  semantic similarity of natural language labels as well as with a visual similarity measure.}
\label{Tab:top_1_CUB}
\begin{tabular}{lcccccccccccc}
{} & {} & {}& {} &{} & {} &{} &\multicolumn{3}{c}{$\boldsymbol{\epsilon}$} & \multicolumn{3}{c}{$\boldsymbol{\epsilon}$ visual similarity} \\
\cmidrule{8-13}
{} & \textbf{SL} &            \textbf{LS} &       \textbf{DML}$_{\boldsymbol{1}}$ &  \textbf{DML}$_{\boldsymbol{2}}$ &  \textbf{KD} & \textbf{CN} &             $\boldsymbol{0.9}$ &           $\boldsymbol{0.99}$ &     $\boldsymbol{0.999}$ &             $\boldsymbol{0.9}$ &           $\boldsymbol{0.99}$ &     $\boldsymbol{0.999}$\\
\midrule
\cmidrule{2-13}
&\multicolumn{12}{c}{DR = 100\%}\\
\cmidrule{2-13}
\ResnetES         & 45.96       & 46.03       & 45.19         & 46.21         & 46.17       & 45.92                  & 46.39         & 46.78          & 47.3            & 46.27           & \textbf{47.75}            & 46.84    \\
\ResnetVS & 46.39       & 46.74       & 46.45         & 46.94         & 46.77       & 46.58                  & 47.35         & 47.96          & 47.2            & 48.15           & 48.08            & \textbf{48.17}     \\
\bottomrule
\end{tabular}
}
\end{center}

\end{table}

\begin{table}[ht!]
    \begin{center}{\caption{CUB-200-2011. Top-5 results, averaged over four trials. We used our method both with the proposed  semantic similarity of natural language labels as well as with a visual similarity measure.}
\label{Tab:top_5_CUB}
\begin{tabular}{lcccccccccccc}
{} & {} & {}& {} &{} & {} &{} &\multicolumn{3}{c}{$\boldsymbol{\epsilon}$} & \multicolumn{3}{c}{$\boldsymbol{\epsilon}$ visual similarity} \\
\cmidrule{8-13}
{} & \textbf{SL} &            \textbf{LS} &       \textbf{DML}$_{\boldsymbol{1}}$ &  \textbf{DML}$_{\boldsymbol{2}}$ &  \textbf{KD} & \textbf{CN} &             $\boldsymbol{0.9}$ &           $\boldsymbol{0.99}$ &     $\boldsymbol{0.999}$ &             $\boldsymbol{0.9}$ &           $\boldsymbol{0.99}$ &     $\boldsymbol{0.999}$\\
\midrule
\cmidrule{2-13}
&\multicolumn{12}{c}{DR = 100\%}\\
\cmidrule{2-13}
\ResnetES         & 72.85 &	73.17&	72.66&	72.78 &	73.67 &	72.92 &	73.78 &	73.29 &	73.19 &	\textbf{75.14} &	73.45 &	73.11
   \\
\ResnetVS & 74.47 &	74.86 &	74.55 &	74.88 &	74.83 & 	74.55 &	\textbf{76.52} &	75.71 &	75.64 &	75.76 &	75.97 &	75.3
  \\
\bottomrule
\end{tabular}
}
\end{center}

\end{table}

\begin{table}[ht!]
    \begin{center}{\caption{CUB-200-2011. Top-1 results, standard deviations over four trials. We used our method both with the proposed  semantic similarity of natural language labels as well as with a visual similarity measure.}
\label{Tab:top_1_CUB_std}
\begin{tabular}{lcccccccccccc}
{} & {} & {}& {} &{} & {} &{} &\multicolumn{3}{c}{$\boldsymbol{\epsilon}$} & \multicolumn{3}{c}{$\boldsymbol{\epsilon}$ visual similarity} \\
\cmidrule{8-13}
{} & \textbf{SL} &            \textbf{LS} &       \textbf{DML}$_{\boldsymbol{1}}$ &  \textbf{DML}$_{\boldsymbol{2}}$ &  \textbf{KD} & \textbf{CN} &             $\boldsymbol{0.9}$ &           $\boldsymbol{0.99}$ &     $\boldsymbol{0.999}$ &             $\boldsymbol{0.9}$ &           $\boldsymbol{0.99}$ &     $\boldsymbol{0.999}$\\
\midrule
\cmidrule{2-13}
&\multicolumn{12}{c}{DR = 100\%}\\
\cmidrule{2-13}
\ResnetES     &    0.18	& 0.21	& 0.23& 	0.35& 	0.21& 	0.29& 	0.24& 	0.19& 	0.19& 	0.16& 	\textbf{0.16}& 	0.09
  \\
\ResnetVS & 0.14& 	0.43& 	0.19& 	0.18& 	0.43& 	0.32& 	0.18	& 0.18	& 0.16& 	0.12& 	0.17& 	\textbf{0.06}
   \\
\bottomrule
\end{tabular}
}
\end{center}

\end{table}

\begin{table}[ht!]
    \begin{center}{\caption{CUB-200-2011. Top-5 results, standard deviations over four trials. We used our method both with the proposed  semantic similarity of natural language labels as well as with a visual similarity measure.}
\label{Tab:top_5_CUB_std}
\begin{tabular}{lcccccccccccc}
{} & {} & {}& {} &{} & {} &{} &\multicolumn{3}{c}{$\boldsymbol{\epsilon}$} & \multicolumn{3}{c}{$\boldsymbol{\epsilon}$ visual similarity} \\
\cmidrule{8-13}
{} & \textbf{SL} &            \textbf{LS} &       \textbf{DML}$_{\boldsymbol{1}}$ &  \textbf{DML}$_{\boldsymbol{2}}$ &  \textbf{KD} & \textbf{CN} &             $\boldsymbol{0.9}$ &           $\boldsymbol{0.99}$ &     $\boldsymbol{0.999}$ &             $\boldsymbol{0.9}$ &           $\boldsymbol{0.99}$ &     $\boldsymbol{0.999}$\\
\midrule
\cmidrule{2-13}
&\multicolumn{12}{c}{DR = 100\%}\\
\cmidrule{2-13}
\ResnetES     &   0.27&	0.42&	0.43&	0.28&	0.22&	0.61&	0.21&	0.23&	0.19&	\textbf{0.15}&	0.29&	0.23

  \\
\ResnetVS & 0.41	0.19&	0.15&&	0.27&	0.27&	0.54&	\textbf{0.19}&	0.31&	0.18&	0.24	&0.37&	0.15

   \\
\bottomrule
\end{tabular}
}
\end{center}

\end{table}

\end{document}